\documentclass[letterpaper,journal]{IEEEtran}

\usepackage{amsthm}
\usepackage{booktabs}

\usepackage{multirow}
\usepackage{lscape} 
\usepackage{tabularx}
\usepackage{comment}
\usepackage{graphicx}
\usepackage{color}
\usepackage{amssymb,amsmath}

\usepackage{xspace}
\usepackage{cite}
\usepackage{hyperref}

\newtheorem{problem}{Problem}

\newtheorem{theorem}{Theorem}
\newtheorem{lemma}{Lemma}
\newtheorem{definition}{Definition}

\newtheorem{remark}{Remark}

\usepackage[linesnumbered,ruled,vlined]{algorithm2e}

\title{HEHA: Bounded Suboptimal Planning for Constrained Routing and Its Use in Heterogeneous Multi-Robot Exploration of Unknown Environments}

\author{Longrui~Yang$^{1,2,*}$,~\IEEEmembership{Student Member,~IEEE,}
        Yiyu~Wang$^{1,3,*}$,~\IEEEmembership{Student Member,~IEEE,} \\
        Jingfan~Tang$^{1,*}$,~\IEEEmembership{Student Member,~IEEE,}
        Yunpeng~Lyu$^{1}$,
        Shizhe~Zhao$^{1}$,~\IEEEmembership{Member,~IEEE,} \\
        Chao~Cao$^{4}$,~\IEEEmembership{Member,~IEEE,} and~Zhongqiang~Ren$^{1,\dagger}$,~\IEEEmembership{Member,~IEEE}%
\thanks{
This work was supported by the Natural Science Foundation of China under Grant 62403313 and the Natural Science Foundation of Shanghai under Grant 24ZR1435900.
\emph{$^{*}$Longrui Yang, Yiyu Wang, and Jingfan Tang contributed equally to this work. $^{\dagger}$Corresponding author: Zhongqiang Ren (email: zhongqiang.ren@sjtu.edu.cn).}}%
\thanks{$^{1}$Longrui Yang, Yiyu Wang, Jingfan Tang, Yunpeng Lyu, Shizhe Zhao, and Zhongqiang Ren are with the Global College, Shanghai Jiao Tong University, Shanghai 200240, China (email: tang\_jingfan@sjtu.edu.cn; lv\_hn\_promising@sjtu.edu.cn; shizhe.zhao@sjtu.edu.cn; zhongqiang.ren@sjtu.edu.cn).}%
\thanks{$^{2}$Longrui Yang is also with Nanyang Technological University, Singapore 639798 (email: longrui001@e.ntu.edu.sg). This work was done while Longrui Yang was a research assistant with Shanghai Jiao Tong University.}%
\thanks{$^{3}$Yiyu Wang is also with the University of Michigan, Ann Arbor, MI 48109, USA (email: wangyiyu@umich.edu).}%
\thanks{$^{4}$Chao Cao is with the Robotics Institute, Carnegie Mellon University, Pittsburgh, PA 15213, USA (email: ccao1@andrew.cmu.edu).}}

\begin{document}

\newcommand\abbrName{HEHA\xspace}
\newcommand\abbrOurMTSP{RCH\xspace}

\maketitle

\begin{abstract}
This paper considers the path planning problem for autonomous exploration of an unknown environment using multiple heterogeneous robots such as drones, wheeled, and legged robots, with different capabilities to traverse complex terrains.
A key challenge there is to intelligently allocate the robots to the unknown areas to be explored and determine the visiting order of those areas subject to traversability constraints, which leads to a large-scale constrained multiple hamiltonian path problem (mHPP) that needs to be quickly and iteratively solved every time when new space is explored.
To address the challenge, we propose \abbrName (Hierarchical Exploration with Heterogeneous Agents) using a recent hierarchical method that decomposes the exploration into global planning and local planning.
The major contribution in \abbrName is its global planning, where we propose a new routing algorithm \abbrOurMTSP (Routing with Constraints and Heterogeneity) that can quickly find bounded sub-optimal solutions for mHPP to minimize the maximum path length (i.e., makespan) among the agents subject to traversability constraints.
Additionally, the local planner in \abbrName also considers heterogeneity to avoid repeated and duplicated exploration among the robots.
The experimental results show, in the presence of heterogeneous traversability, our \abbrOurMTSP can solve mHPP with up to 47\% smaller makespan than several baselines, and \abbrName can reduce the exploration time by up to 30\% compared with existing approaches.

\end{abstract}

{\begin{IEEEkeywords}
Multi-robot systems, Path planning, Traveling salesman problems, Autonomous exploration
\end{IEEEkeywords}}

\section{Introduction}
\label{sec:intro}

\begin{figure}[tb]
    \centering
    \includegraphics[width=0.92\linewidth]{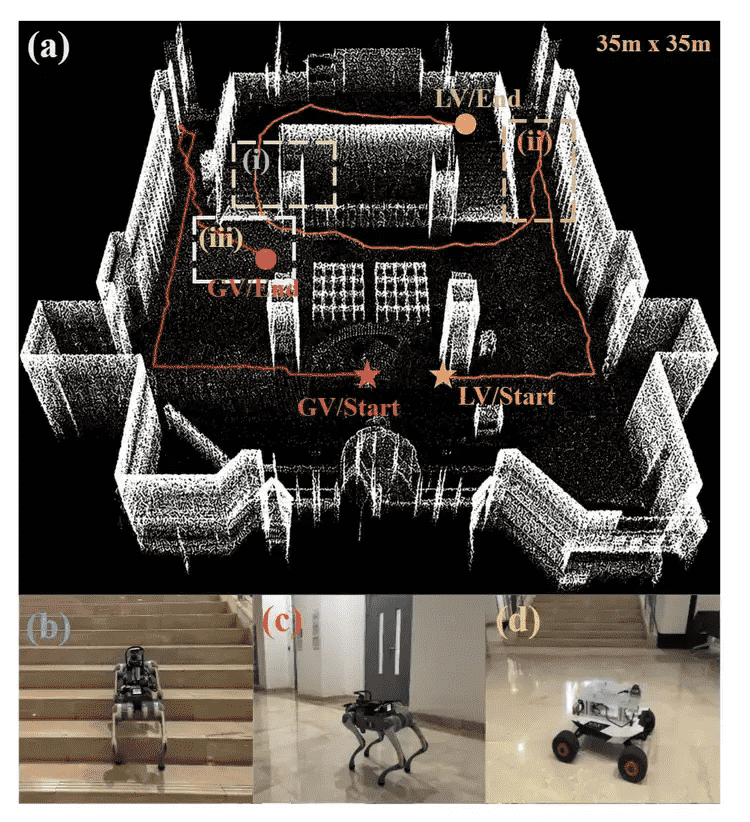}
    \caption{Exploration using a ground vehicle (GV) and a legged vehicle (LV).
    (a) shows the results after exploration where the orange and red lines show the robots' trajectories. (b-d) are images of the real world for the areas (i-iii) in (a). The stairs in area (i) as shown in (b) are first found by GV but is only accessible by the LV. Our planner considers such traversability constraints during path planning. Our video attachment shows the exploration process.}
    \label{fig:exploration results}
\end{figure}

Autonomous exploration seeks to navigate one or multiple robots in a bounded unknown environment so as to map the environment as soon as possible, which is of fundamental importance in robotics and arises in applications such as underground mining~\cite{underground_mining} and planetary exploration~\cite{planetary_exploration}.
This paper considers autonomous exploration using a team of heterogeneous robots such as drones, wheeled, and legged robots, which have different capabilities to traverse complex terrains.
For example, wheeled robots cannot pass stairs while drones and legged robots can do so (Fig.~\ref{fig:exploration results}).
In particular, we consider that all robots always stay in communication, and focus on the planning aspects of the problem: how to quickly plan paths for the robots to minimize the exploration time while considering their heterogeneous terrain traversability.

Autonomous exploration involves various challenges such as mapping~\cite{improve_map,mui_tare}, planning~\cite{tare,frontier_based} and coordination~\cite{racer,comm3}.
While exploration problems have been extensively investigated recently, little of the existing work considers heterogeneous traversability.
The major challenge is that the planner has to consider the allocation of robots to the unknown areas to be explored as well as the visiting order of those areas subject to traversability constraints, which leads to a large-scale constrained optimization problem that needs to be quickly and iteratively solved every time when new space is explored.

To address the challenge, we propose \abbrName, a robotic system for Hierarchical Exploration with Heterogeneous Agents, which builds upon the common frontier-based approaches~\cite{multi_frontier_based} and leverages a recent hierarchical framework~\cite{tare}, with the major difference in its planning components.
Specifically, frontiers are the boundaries between explored and un-explored spaces, and the planner needs to iteratively plan paths for the robots during exploration until the frontiers deplete.
The recent hierarchical framework~\cite{tare} decomposes the problem into global and local planning:
 With full global communication, the global planning divides and groups frontiers into clusters, and then formulates and solves a multiple Hamiltonian Path Problem (mHPP) to allocate to and specify the visiting order of these clusters of frontiers for each robot, where each node in the mHPP graph corresponds to a cluster in exploration.
The local planning then finds a detailed local path for each robot subject to kinodynamic constraints to visit all frontiers in the assigned clusters for exploration.

Within \abbrName, the \textbf{first} and major technical contribution is its global planning, where we propose a new framework for mHPP called \abbrOurMTSP (Routing with Constraints and Heterogeneity) that can quickly return bounded sub-optimal solutions to minimize the maximum path length (makespan) among the robots, while satisfying traversability constraints that are described as node-robot \emph{assignment constraints} in the mHPP graph.
Although mHPP is a well-known NP-hard problem that has been researched a lot, most of the existing work does not consider assignment constraints, and focuses on minimizing the sum of path lengths (min-sum) of the robots as opposed to minimizing the maximum (min-max) as \abbrOurMTSP does.

To name a few, to find solutions with optimality or bounded sub-optimality guarantees, integer programming methods for traveling salesmen problems (TSP) and vehicle routing problems (VRP) usually optimize the min-sum objective and are limited to small-scale problems~\cite{laporte1992vehicle,philip2025mixed}.
Approximation methods~\cite{approx_2_depot_hhpp,approx_k_depot_hpp} convert the problem to simpler ones (such as minimum spanning trees) and then build solutions to the original problem with post-processing in polynomial time.
However, these methods depend on specific structure of the problem and cannot be easily extended to handle assignment constraints or min-max objectives.
Heuristic methods build an initial solution and then iteratively refine this solution, which often run fast and can be adapted to handle various constraints, yet lack solution quality guarantees~\cite{k_opt_lkh,ortools_routing,pyvrp}.

In exploration and other applications~\cite{tare,DMS} with the need of ``real-time'' planning with limited on-board computational resources, solvers with heuristic algorithms such as PyVRP~\cite{pyvrp} and Google OR-Tools~\cite{ortools_routing} are often applied.
However, both of them are designed for min-sum objectives.
Simply applying them to handle the min-max problem with assignment constraints as required in this work, can either lead to infeasible paths due to the omission of the assignment constraint or longer exploration time due to the min-sum optimization.
Fundamentally, the mHPP suffers from the curse of dimensionality which grows exponentially as the number of robots and graph nodes increases.

To address the challenge, we propose \abbrOurMTSP, a graph search framework with completeness and solution bounded sub-optimality guarantees, which is fundamentally different from existing TSP or VRP methods.
Conventionally, for TSP and VRP problems, search methods suffer from a large branching factor and are under-explored.
We take the view that by designing a framework that can fuse existing search techniques, which are often studied in other problem domains very different from TSP/VRP, in a principled and novel way, the mHPP problem can be potentially quickly solved with solution quality guarantees.

With that in mind, our \abbrOurMTSP runs systematic A*-like search over the state space to incrementally construct paths, as guided by new heuristic functions tailored for the problem which relax and provide lower bounds to the remaining problems, while using new dominance rules to prune non-promising partial solutions to save computational effort.
During the search, after obtaining each complete solution, that solution is guaranteed to be bounded sub-optimal, and \abbrOurMTSP further leverages a local refinement procedure to post-optimize the solution quality.
The systematic search in \abbrOurMTSP may also help avoid being trapped into a local minimum after local optimization.
We regard \abbrOurMTSP as a framework since different node selection and pruning criteria can be employed during the search, various relaxation can be used for heuristic functions, and different local refinement methods can be used for post-optimization.

The \textbf{second} technical contribution in \abbrName is its local planner, which develops two simple tricks to include heterogeneity into local planning by prioritizing robots of different types based on their traversability.
To maintain fast running speed, the local planner in \abbrName incorporates the heterogeneity as cost terms when planning detailed motion of the robots to avoid repeated and duplicated exploration among the robots.
Such simple rules allow fast and reactive adaptation of robots' motion based on heterogeneity in local areas.

We test our \abbrOurMTSP for mHPP, and \abbrName for exploration.
First, we compare our \abbrOurMTSP against several baselines including PyVRP~\cite{pyvrp} and Google OR-Tools~\cite{ortools_routing} and their fine-tuned versions, on various mHPP instances with assignment constraints, and our \abbrOurMTSP finds solutions with up to 47\% smaller makespan than the baselines.
Second, we compare our \abbrName on three maps in simulation against two baselines that adapt the popular next-best-view planner (NBVP)~\cite{nbvp} and a greedy method to heterogeneous robots,  with full global communication.
The results show that \abbrName requires up to 30.2\% less exploration time and up to 35.7\% shorter total path length than the baselines.
Finally, we demonstrate \abbrName on physical robots, including a wheeled robot and a legged robot in a bounded space of size about $35m\times 35m$ with flat ground and stairs, and \abbrName is able to intelligently navigate the legged robot to explore the stair regions while letting the wheeled robot explore the vast flat ground areas to minimize the completion time of exploration.

\section{Related Work}
\label{sec:related_work}

\subsection{Autonomous Exploration}

Single-robot exploration was studied using frontier-based~\cite{multi_frontier_based,frontiers_base}, information theoretic approaches~\cite{inf_computationally,information_bayesian_opt}, receding horizon sampling~\cite{nbvp}, traveling salesman methods~\cite{tare,tare_planner} and learning~\cite{drl_office,drl,drl_large_scale}, to name a few.
Multi-robot exploration is usually more challenging than single-robot due to the coordination and interaction among the robots, such as communication limitation~\cite{comm,comm3,imec,comm2,sim2real_multi_exploration}, map merging across the robots~\cite{mui_tare}, assignment and ordering the areas to be explored among the robots~\cite{gbp}, etc.

Most existing multi-robot exploration considers homogeneous robots, such as wheeled~\cite{cure}, legged \cite{mine_tunnel_dog}, or flying robots~\cite{racer}.
There are only a few works on heterogeneous robots, with the focus on using flying robots for coarse survey followed by detailed exploration using ground vehicles~\cite{dronefly_first,improve_map,aage}, or routing with heterogeneous rewards from various sensors on different robots~\cite{hetero_teams_exploration}.
Although some work builds systems leveraging the traversability of different robots in complex environments~\cite{gbp,hetero_teams_exploration,csiro_darpa,coped,9705621}, none of them focuses on the planning with heterogeneous traversability.

This work follows the general hierarchical framework for autonomous exploration in~\cite{tare} where the global planning decomposes the space into regions and determines the visiting order of those regions by solving a routing problem, while the local planning finds a detailed trajectory for the robot to explore each of those regions.
When applying this hierarchical method to heterogeneous robots, a major difficulty lies in its global planning, which requires quickly solving an mHPP problem with assignment constraints, and is addressed by this work.
We do not consider limited communication among the robots and assume global communication during exploration.

\subsection{Routing Problems with Multiple Robots}
\label{sec:related:mHPP}
For multi-robot exploration \cite{racer,mui_tare} and environment reconstruction \cite{multi_scene_reconstruction}, the coordination of robots is often achieved by iteratively solving mHPP and VRP problems, which are well-known NP-hard problems \cite{TSP}.
These routing problems have been approached by exact methods \cite{laporte1992vehicle,today_TSP}, bounded sub-optimal approximations \cite{approx_2_depot_hhpp,approx_k_depot_hpp}, and unbounded sub-optimal heuristics \cite{k_opt_lkh,ortools_routing,pyvrp}, trading off solution quality for runtime efficiency. 
However, most existing routing algorithms optimize the min-sum objective~\cite{approx_k_depot_hpp,approx_2_depot_hhpp,11237208,today_TSP} to reduce total energy or distance.
While sometimes suitable for applications in logistics~\cite{cbss,DMS}, min-sum formulations can lead to long makespan and thus prolong the exploration.

Despite these progresses, few of them consider assignment constraints~\cite{approx_2_depot_hhpp,algo_hvrp} that are necessary to describe heterogeneous terrain traversability of robots in exploration.
In particular, transformation methods~\cite{today_TSP,cbss} can handle assignment constraints, but they can only minimize the sum of costs, which can lead to long makespan.

Currently, robotic exploration systems often rely on public routing solvers such as Google OR-Tools~\cite{ortools_routing}, PyVRP~\cite{pyvrp} and LKH~\cite{k_opt_lkh}, which are usually designed for min-sum objectives originally.
Although some of them provide options and parameters to configure the solver to indirectly minimize makespan subject to assignment constraints, they usually run slowly or return infeasible or poor quality solutions that still violate the assignment constraints, which eventually leads to poor performance in robotics application.

\section{Preliminaries}

\subsection{Problem Statement of Exploration} \label{sec:problem}

Let $\mathcal{W} \subseteq \mathbb{R}^3$ denote a bounded 3D space to be explored, and let $\mathcal{S} \subset \mathcal{W}$ denote all the surfaces in the workspace to be mapped, which is the boundary between obstacles $\mathcal{W}_{\text{obs}}$ and free space $\mathcal{W}_{\text{free}}$.
Let $I=\{1,2,\cdots,N_a\}$ denote a set of $N_a$ robots, and we use a superscript $i$ over a variable to indicate to which robot that variable belongs.
Let $q^i\in \mathrm{SE}(3)$ denote the view pose of the on-board sensor on robot $i\in I$, which can perceive a volume $\mathcal{W}(q^i)\subseteq \mathcal{W}_{\text{free}}$ that is a subset of the obstacle-free workspace.
Let $\mathcal{S}(q^i)\subseteq \mathcal{S}$ denote the surface that the robot perceives at pose $q^i$.

Let $A: \mathrm{SE}(3) \rightarrow 2^I$ denote the \emph{capability map}, which maps any pose $q\in \mathrm{SE}(3)$ to a subset of robots $A(q)\subseteq I$ that can reach that pose $q$.
Here, $2^I$ is the power set of $I$.
This paper assumes $A$ is available for any pose within the perceived volume of any robot during exploration.
In practice, additional terrain analysis is needed to get $A$.

All robots share the same global clock and start their motion at time $t=0$.
Let $\tau^i=\{q^i_1,q^i_2,\cdots,q^i_\ell\}$ denote a trajectory of robot $i$, where each trajectory point is an element of $\mathrm{SE}(3)$, and any two subsequent trajectory points have a time interval $t_\Delta$ in between.
A trajectory is \emph{feasible} if it satisfies both the kinodynamic constraints of the robot and \emph{traversability constraints}, that is, $i \in A(q^i_k), k=1,2,\cdots,\ell$.
Let $\operatorname{cost}(\tau^i)=(\ell-1) t_\Delta$ denote the finish time of trajectory $\tau^i$, and let $\bigcup_{q^i\in \tau^i}\mathcal{S}(q^i)$ denote the union of all surfaces that are perceived by the robots along $\tau^i$.

\begin{problem}[Exploration]
The goal of the exploration problem considered in this paper is to find a set of trajectories $\{\tau^i,i\in I\}$ such that (i) each trajectory $\tau^i$ is feasible, (ii) all surfaces are perceived $\bigcup_{q^i\in \tau^i, i\in I}\mathcal{S}(q^i) = \mathcal{S}$, and (iii) the exploration time $\max_{i\in I}\operatorname{cost}(\tau^i)$ reaches the minimum.
\end{problem}

\subsection{System Overview}

\begin{figure*}[t]
    \centering
    \includegraphics[width=\linewidth]{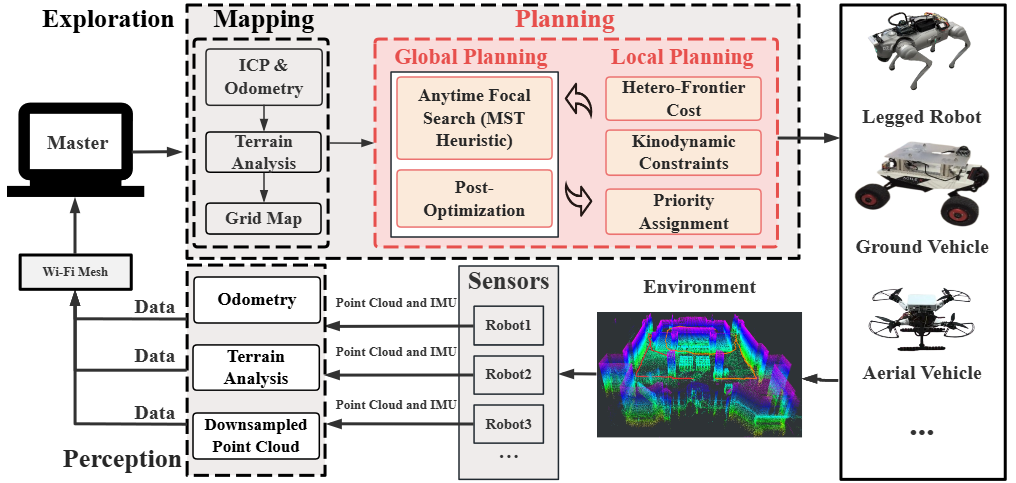} 
    \caption{\abbrName System Overview. The red boxed modules, namely global and local planning, are the innovative part of our approach. Other components are similar to the existing hierarchical exploration system~\cite{tare}. Different types of robots collect point clouds, and IMU measurements through onboard sensors. Then each robot transmits them back to the master computer via the communication network. The system first performs Iterative Closest Point based registration and combines the results with terrain analysis to construct a grid map. Finally, the planning layer conducts global and local path planning.}
    \label{fig:p1}
\end{figure*}

Our \abbrName system follows the same architecture as in~\cite{tare}, which consists of Perception, Global Planning, Local Planning, and Motion Control (Fig.~\ref{fig:p1}).
The major difference lies in the planning modules including both the global and local planning.
All robots are planned and managed in a centralized fashion in a sense that there is global communication among the robots.

The perception module runs Lidar SLAM to both build a point cloud map of the surrounding environment and estimate the pose of the robot.
All robots start from initial poses with known relative transformation so that all robots' point cloud maps can be transformed to the same common frame.
The perception also runs terrain analysis to understand the areas that each robot can access, as elaborated in Sec.~\ref{sec:terrain}.

During exploration, the boundary between known and unknown areas forms the frontiers and these frontiers are to be visited by the robots to explore the environment.
These frontiers are grouped into clusters, and the global planning module determines the visit order of these clusters by solving a routing problem on a graph, in which each graph vertex corresponds to a cluster of frontiers, and each graph edge corresponds to the traversal between two clusters with an estimated traversal cost.
Sec.~\ref{sec:global} presents the global planning.

Given the visit order of clusters, the local planning module (Sec.~\ref{sec:local}) plans a detailed trajectory in the region enclosing the frontiers of that cluster, for exploration subject to the robots' kinodynamic constraints.
The planned trajectory is then sent to the motion control for execution.
The entire process iterates until the environment is fully explored.
We detail the local and global planning later in Sec.~\ref{sec:local_global_interaction}

\section{Global Planning}
\label{sec:global}

\subsection{Problem Statement of Global Planning}

\begin{figure}[t]
    \centering
    \includegraphics[width=\linewidth]{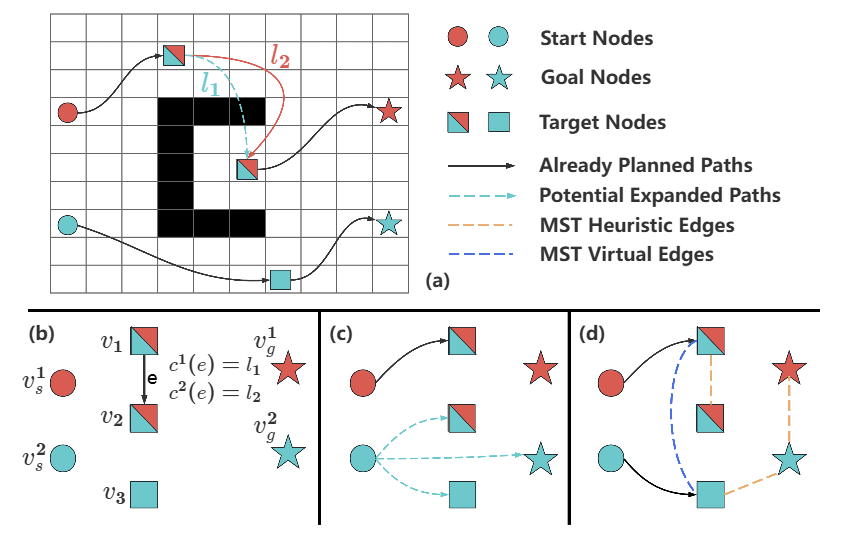} 
    \caption{(a) An example of global planning, where a path in the workspace for a robot corresponds to an edge in the target graph in (b), and the edge cost in (b) is same as the corresponding path length in (a). The color of a target node indicates the assignment constraints. The objective is to find a joint path in the mHPP graph, as in (b), for all robots that minimizes the maximum path cost. (b) The corresponding complete target graph, where $v_s^i$, $v_g^i$ are the start and goal nodes of robots $i=1,2$ and $v_1,v_2,v_3$ are the target nodes to be visited. (c) Visualization of a search label, which represents a partial solution and its possible successors during expansion. Here, to expand this label, the next moves of robot $2$ are considered as shown by the dashed lines, which correspond to a feasible target or goal node.
    (d) The proposed MST-based heuristic estimates a lower bound of the remaining cost to visit all unvisited targets, where current robot positions are connected by zero-cost virtual edges.}
    \label{global}
\end{figure}

Let $G=(V,E)$ denote a complete graph, where $V=\{v_1,v_2,...,v_{N_v}\}$ is the set of $N_v$ nodes, and $E$ is the set of edges connecting all pairs of nodes.
Each node $v\in V$ is associated with a set of agents $A(v)\subseteq I$ that is capable of visiting the node $v$.\footnote{
Here we abuse the notation $A$, which is previously used to denote the capability map in the problem definition in Sec.~\ref{sec:problem}. Both places share the common meaning that $A$ maps either a pose $q\in \mathrm{SE}(3)$ or a node $v\in G$ to a subset of agents that can visit that pose or node.}
These sets $A(v), v\in V$ specify the node-agent assignment constraints in $G$ (Fig.~\ref{global}(a)).
The cost between any two nodes $u,v\in V$ when traversed by robot $i$ is denoted as $c^i(u,v)$.
If robot $i$ cannot reach either $u$ or $v$, then $c^i(u,v)$ is infinity.
In $G$, each agent $i$ has a start node $v^i_s$ and a goal node $v^i_g$.
Each agent $i$ seeks a path $\pi^i$, an ordered sequence of nodes in $G$.
The joint path $\pi = \{\pi^1,\pi^2,...,\pi^k\}$ is the set of all agents' paths.
Let $C(\pi^i)$ denote the cost of path $\pi_i$, which is $C(\pi^i)=\sum_{(v_k,v_{k+1}) \in \pi^i}c^i(v_k,v_{k+1})$.

\begin{problem}[Min-Max mHPP]\label{heha:problem:mhpp}
The Min-Max Multiple Hamiltonian Path Problem (mHPP) seeks a joint path of all agents while minimizing $\max_{i \in I} C(\pi^i)$, the maximum path cost among all agents (i.e. makespan).
\end{problem}

\begin{remark}
In exploration, each node in $G$ corresponds to a frontier $q\in \mathrm{SE}(3)$ to be visited, and an edge indicates the transition from one frontier to another. The edge cost $c^i(u,v)$ is estimated by solving a small path planning problem from $u$ to $v$ based on the kinodynamic constraints of the robot $i$.
 The min-max objective in Problem~2 is aligned with the exploration objective in Problem~1: since all robots explore simultaneously, the exploration is completed when the last robot finishes.
Therefore, minimizing the maximum routing cost provides the global planner with an objective directly related to the overall exploration completion time.
\end{remark}

\subsection{Integer Programming Formulation}
Similar to many other routing problems, the Min-Max mHPP considered in this work can also be formulated as an integer program (IP).
Let binary variables $x^i_{uv} \in \{0,1\}$ indicate whether robot $i$ traverses the edge $(u,v)$.
Let $V_T := V \setminus \{s_i, g_i : i \in I\}$ denote the set of target nodes to be visited, and let $N_T = |V_T|$.
Let $u_v \in \{1, \dots, N_T\}$, $v \in V_T$, be auxiliary variables that indicate the visiting order of the target nodes.
Problem \ref{heha:problem:mhpp} can be formulated as follows.
\begin{align}
\min \quad & \max_{i\in I} \left(\sum_{u,v \in V} c^i(u,v) x^i_{uv}\right) \label{mip:obj}\\
\text{s.t.} \quad
& \sum_{i \in I} \sum_{u \in V} x^i_{uv} = 1, \quad \forall v \in V_T \label{mip:cstr:flow1}\\
& \sum_{v \in V} x^i_{s_i v} = 1, \quad \sum_{u \in V} x^i_{u g_i} = 1, \quad \forall i \in I \label{mip:cstr:start_end}\\
& \sum_{u \in V} x^i_{u s_j} = 0, \quad \sum_{v \in V} x^i_{g_j v} = 0, \quad \forall i, j \in I \label{mip:cstr:no_in_start_no_out_end}\\
& \sum_{v \in V} x^i_{uv} = \sum_{v \in V} x^i_{vu},
\quad \forall i \in I, u \in V_T \label{mip:cstr:conservation}\\
& x^i_{uv} = 0, \quad \text{if } i \notin A(u) \text{ or } i \notin A(v) \label{mip:cstr:hetero}\\
& u_w - u_v + N_T \sum_{i \in I} x^i_{vw} \le N_T - 1,
\,\forall v, w \in V_T, v \neq w \label{mip:cstr:mtz}\\
& x^i_{uv} \in \{0,1\}, \quad u_v \in \{1, \dots, N_T\} \label{mip:cstr:decision}
\end{align}
Constraint \ref{mip:cstr:flow1} ensures each target vertex is visited exactly once by exactly one robot.
Constraint \ref{mip:cstr:start_end} ensures each robot must start from its own starting point and end at its own ending point. 
Constraints \ref{mip:cstr:no_in_start_no_out_end} prevents any robot from entering a starting point or leaving a goal point, respectively.
Constraint \ref{mip:cstr:conservation} guarantees that if a robot enters an intermediate vertex, it must also depart from it.
Constraint \ref{mip:cstr:hetero} enforces the node-agent assignment constraints so that a robot $i$ cannot traverse an edge incident to a vertex that it is not allowed to visit.
Constraint \ref{mip:cstr:mtz} eliminates disconnected cycles.
Constraint \ref{mip:cstr:decision} states that the edge variables are binary and the order variables are integers.
Finally, the objective function seeks to minimize the maximum path costs of all agents and each agent's path cost is the sum of edge costs along its path.

\subsection{\abbrOurMTSP Overview}

To solve Problem~\ref{heha:problem:mhpp}, \abbrOurMTSP runs a systematic A*-like search that incrementally extends partial solutions (joint paths), one vertex at a time, until all vertices in $V$ are visited and all robots reach their goals.
A direct implementation of this search is intractable, since the branching factor grows exponentially in the number of robots.

\abbrOurMTSP therefore combines five components. 
Partial expansion (Sec.~\ref{sec:global:seq_exp}) advances a single robot per expansion, keeping the branching factor independent of $N_a$.
A dominance rule (Sec.~\ref{sec:concepts}) prunes partial solutions that cannot lead to a better makespan than an already generated one. 
An MST-based heuristic (Sec.~\ref{sec:heu}) lower bounds the makespan of any solution extending a given partial solution, and a focal list (Sec.~\ref{sec:focal}) selects among the remaining partial solutions those closest to completion, within a user-specified suboptimality bound. 
Finally, a post-optimization (Sec.~\ref{sec:post-opt}) procedure refines every solution found during the search.

Among these components, partial expansion~\cite{partial_expansion}, focal search~\cite{A*_study} and 2-opt~\cite{k_opt_lkh} are adapted from the literature, where they were developed for problem domains other than mHPP. 
The components that are new in RCH are (i) the heuristic, which lower bounds the cost-to-go of a partial solution for min-max mHPP under assignment constraints; (ii) the dominance rule, which compares partial solutions by both the cost of each robot and the set of visited vertices; and (iii) the post-optimization, which perturbs the paths within and across groups of robots of the same type.

\subsection{Concepts and Notations}
\label{sec:concepts}

Let $\mathcal{G}=G^{N_a}=G\times G\times \cdots \times G$ denote the \emph{joint graph} of all robots.
Our \abbrOurMTSP uses A*-like search to iteratively plan paths in $\mathcal{G}$ from the joint start vertex $\vec{v}_s =(v^1_s,v^2_s,\cdots,v_s^{N_a})$ of all robots towards their joint goals $\vec{v}_g = (v^1_g,v^2_g,\cdots,v^{N_a}_g)$.

\subsubsection{Label and Dominance}
During the search, let $l=(\vec v, \vec g, B)$ denote a \textit{label}, where 
\begin{itemize}
    \item $\vec v = (v^1,v^2,\cdots,v^{N_a})$ is a joint vertex that represents the current vertices of the robots in this label, 
    \item the cost vector $\vec g=(g^1,g^2,\cdots,g^{N_a})$ denotes the current cumulative costs of the robots, and 
    \item the set $B\subseteq V$ is the set of nodes that have been visited.
\end{itemize}
During planning, each label identifies a unique path in $\mathcal{G}$ that can be reconstructed by iteratively backtracking the \emph{parent pointers} until reaching $l_0$.
We use $\vec{v}(l),\vec{g}(l),B(l)$ to denote the corresponding component in label $l$.
Let $l_0=(\vec{v}_s,\vec{0},\emptyset)$ denote the initial label.

There can be multiple paths in $\mathcal{G}$ from $v_s$ to any other $v\in G$, and the planner needs to compare them and potentially discard a less promising one.
We introduce the following dominance rule to compare and prune labels, which is similar to some recent work that uses A*-like search to solve (single-agent) TSP or HPP problems~\cite{RMA,2025_SOCS_BOTSPTW_ShizheZhao}.
\begin{definition}[Label Dominance]\label{heha:def:dom}
Given two labels $l_1=(\vec{v}_1,\vec{g}_1,B_1)$ and $l_2=(\vec{v}_2,\vec{g}_2, B_2)$, we say that $l_1$ \textit{dominates} $l_2$, denoted $l_1 \prec l_2$, if the following conditions hold: (i) $\vec v_1 = \vec v_2$; (ii) $\vec{g}_1$ is element-wise no greater than $\vec{g}_2$, which is denoted as $\vec g_1 \le \vec g_2$; (iii) $B_2 \subseteq B_1$. We also write $l_1\preceq l_2$ if $l_1 \prec l_2$ and $l_1=l_2$.
\end{definition}
If label $l_1\preceq l_2$, then $l_2$ cannot lead to a better solution path from $v_s$ to $v_g$ since any future path from $l_2$ can be cut-and-paste to $l_1$ without worsening the makespan.
We prove the correctness of this dominance rule in Lemma~\ref{heha:lem:dom} in Sec.~\ref{heha:sec:analysis}.

For two labels $l_1,l_2$ with $\vec{v}(l_1)=\vec{v}(l_2)$, if neither of them dominates the other, these two labels are non-dominated by each other.
At each joint vertex $\vec{v}\in \mathcal{G}$, there can be multiple labels that are non-dominated by each other.
To store these non-dominated labels, let $L(\vec{v})$ denote a \emph{frontier set} of non-dominated labels with $\vec{v}(l)=\vec{v}$.
During planning, when a new label $l'$ is generated, $l'$ is checked for dominance against existing labels in $L(\vec{v}(l'))$ as explained later.

\subsection{Partial Expansion}\label{sec:global:seq_exp}
Each vertex in the joint graph $\mathcal{G}$ has $(|V|-1)^{N_a}$ successor vertices, and this number is referred to as the branching factor of the search.
Naively expanding a label $l$ by generating all labels for those successor vertices is computationally prohibitive as the branching factor grows exponentially
in $N_a$.
To bypass this curse of dimensionality, our \abbrOurMTSP thus takes a partial expansion strategy~\cite{partial_expansion}, which has been used in Multi-Agent Path Finding problems but not yet been investigated under the mTSP or mHPP settings.
Intuitively, partial expansion here plans the next actions of the robots one by one during the search, as opposed to planning a next joint action of all robots together, which helps reduce the branching factor.
Such reduction of branching factor comes at the cost of moving one agent each time as opposed to move multiple agents each time in $\mathcal{G}$, which therefore leads to more iterations of search, which is addressed by dominance pruning and focal search as explained next.

Specifically, for partial expansion, let $A_{\text{active}}(l) \subseteq I$ denote the set of robots that have not yet reached their goal nodes.
In other words, any ``inactive'' robot $j\notin A_{\text{active}}(l)$ must have reached their goal, i.e., $\vec{v}^{\,j}(l) = v^j_g$.
Let $i(l)$ denote the robot in $A_{\text{active}}(l)$ with the smallest cost value in $\vec{g}(l)$.
To expand a label $l$, \abbrOurMTSP only considers the possible movement of robot $i$ from $\vec{v}^i(l)$ to another unvisited vertex $u\in V,u\notin B(l)$, while keeping the vertices of all other robots the same.
For each generated successor label $l'$, the edge cost value traversed by robot $i$ is added to the corresponding component (in $\vec{g}$, $g^i(l') \gets c^i(u,v) + {g}^i(l)$, and the vertex $u$ is marked as visited, i.e., $B(l')\gets B(l)\cup \{u\}$).

\subsection{Heuristics}
\label{sec:heu}

For any label $l$, let $g_{\text{max}}(l)=\max_{i\in I}\vec{g}^{\,i}(l)$ denote the current maximum cost among all robots, which resembles the $g$-value in regular A* search.
To estimate the remaining cost-to-go, we introduce a heuristic function $h(l)$ based on the minimum spanning tree (MST) among all unvisited nodes $V\setminus B(l)$, which is computed with the following steps.

First, for label $l$, the current nodes of all robots $\vec{v}(l)$ are connected with zero-cost virtual edges, and then a MST is computed to connect all unvisited nodes $V\setminus B(l)$. 
With these zero-cost edges, the MST computed as explained next will connect all currently unvisited nodes $V\setminus B(l)$ together with the current robot locations.

Second, when computing this MST, for each edge $e\in E$, the edge cost value is selected to be the minimum among all robots $c(e)=\min_{i\in I}c^i(e)$ so that the total cost of the resulting MST is a lower bound on the cost-to-go. 
If a robot cannot traverse the edge due to assignment constraints, the corresponding cost is infinity.
By doing so, the resulting edge cost represents the most ``optimistic'' traversal cost achievable by any robot.

Third, let $c_{\text{MST}}(l)$ denote the total edge cost of this MST.
Recall that our planner minimizes the maximum path costs of the agents. Then, the cost of this MST is divided by the number of active agents $h(l):=\frac{c_{\text{MST}}(l)}{|A_{\text{active}}(l)|}$, which is a lower bound on the cost-to-go from $l$.

Fourth, to tighten the estimated lower bound, we introduce $f_{\text{est}}(l)=\min_{i\in A_{\text{active}}}(\vec{g}^{\,i}(l) + h(l))$, which estimates the smallest total path cost of any active robots.
$f_{\text{est}}(l)$ is a lower bound on the makespan of robots in $A_{\text{active}}$.

Finally, we define the $f$-value of label $l$ as $f(l)=\max \{g_{\text{max}}(l),f_{\text{est}}(l)\}$, which considers both active and inactive robots.
This $f(l)$ is a lower bound on the makespan of any solution building upon $l$.
We summarize the properties of this MST heuristic with the following lemma.
\begin{lemma}\label{heha:lem:heu_adm}
    The proposed heuristic is admissible: the $f$-value $f(l)=\max \{g_{\text{max}}(l),f_{\text{est}}(l)\}$ of any label $l$ is a lower bound on any solution that can be obtained by extending $l$.
\end{lemma}
\begin{proof}
Consider any feasible complete solution obtained by extending label $l$, and let $M^\star(l)$ denote its makespan. 
For each active robot $i\in A_{\text{active}}(l)$, let $r_i$ denote its remaining cost from label $l$ to its goal in this solution.
As stated before, the MST's total cost $c_{\text{MST}}(l)$ is a lower bound on the total remaining traversal cost of all active robots:
$\sum_{i\in A_{\text{active}}(l)} r_i \ge c_{\text{MST}}(l)$.
It follows that
$$\max_{i\in A_\text{active}(l)} r_i\ge\frac{1}{|A_\text{active}(l)|}\sum_{i\in A_\text{active}(l)} r_i\ge\frac{c_\text{MST}(l)}{|A_\text{active}(l)|}=h(l)$$
Thus, $h(l)$ lower bounds the remaining makespan from $l$.
Let $j\in A_\text{active}(l)$ be an active robot with the maximum remaining cost in the above future expansions, i.e., $r_j=\max_{i\in A_{\text{active}}(l)} r_i$. 
Since $g^j(l)\ge \min_{i\in A_\text{active}(l)} g^i(l)$,
\[
g^j(l)+r_j
\ge
\min_{i\in A_{\text{active}}(l)} g^i(l)+h(l)
=
f_{\text{est}}(l).
\]
The final makespan must be at least the final path cost of robot $j$.
It follows that
$M^\star(l)\ge f_{\text{est}}(l)$.
Moreover, the current maximum cost cannot decrease in any extension, and thus
$M^\star(l)\ge g_{\text{max}}(l)$.
Combining the above inequalities, we obtain
\[
M^\star(l)\ge \max\{g_{\text{max}}(l), f_{\text{est}}(l)\}=f(l).
\]
Hence, $f(l)$ lower bounds the makespan of any feasible solution extending $l$, and the proposed heuristic is admissible.
\end{proof}

\subsection{Focal List}
\label{sec:focal}
\abbrOurMTSP uses focal search to speed up the computation while ensuring bounded sub-optimal solution.
Focal search maintains an open list $L_{O}$ and a focal list $L_{F}$.
$L_{O}$ contains all labels to be expanded, where labels are prioritized based on their $f$-values from the minimum to the maximum.
$L_{F}$ is a subset of $L_{O}$ containing labels with $f$-value within $[f_{\text{min}},(1+\epsilon)f_{\text{min}}]$, where $f_{\text{min}}$ is the smallest $f$-value of labels in $L_{O}$.
$L_{F}$ prioritizes labels based on another criterion as explained next, and the search also selects the label with the highest priority in $L_{F}$ for expansion.
$L_{F}$ prioritizes labels based on the following three values in lexicographic order:
\begin{itemize}
    \item (i) the number of visited nodes $|B(l)|$,
    \item (ii) the $f$-value $f(l)$, and
    \item (iii) $g_{\text{sum}}(l):=\sum_{i\in I}\vec{g}^{\,i}(l)$ the total path cost of all robots in label $l$.
\end{itemize}
In other words, for any two labels $l_1$ and $l_2$, $l_1$ has higher priority than $l_2$ in $L_{F}$, if (i) $|B(l_1)| > |B(l_2)|$, or (ii) $|B(l_1)| = |B(l_2)|$ and $f(l_1) < f(l_2)$, or (iii) $|B(l_1)| = |B(l_2)|$, $f(l_1) = f(l_2)$ and $g_{\text{sum}}(l_1) < g_{\text{sum}}(l_2)$.
If all three values are equal, ties are broken arbitrarily.
Such a focal prefers earlier expansion of labels with more visited nodes.

\begin{algorithm}[t]\label{alg:saf}
\small
\caption{\texttt{\abbrOurMTSP}}
Initialize $l_0$ and compute $f_0$\;
Add $l_0$ to $L_{O}$ and $L_{F}$ // $L_O,L_F$ are open and focal lists\;

\While{within time limit}{
  \While{$L_{F}$ is not empty}{
    $l \gets L_{F}.pop()$\;
    Remove $l$ from $L_{O}$ and $L_{F}$\;
    \If{\texttt{IsDom}($l$)}{\textbf{continue}}
    \texttt{UpdateAndFilter}($l$)

    \If{\texttt{IsCompleteAndBetter}($l$)}{
      $\pi \gets \texttt{PostOptimize}(l)$\;
      \textbf{break}\;
    }
    
    \ForEach{$l'$ in \texttt{SeqExpand($l$)}}{
      \If{\texttt{NotFeasible}($l'$) or \texttt{IsDom }($l'$)}{
        \textbf{Continue}\;
      }
      
        parent($l'$)$\gets l$
      
        Add $l'$ to $L_{O}$ and update $L_{F}$
        
    }
  }
  Decrease $\epsilon$\;
}
\Return $\pi$
\end{algorithm}

\begin{algorithm}[t]\label{alg:postOpt}
\small
\caption{\texttt{PostOptimize}($l$)}
$pathgroups \gets \texttt{SortPathGroup}(l)$\;
\ForEach{group $\in pathgroups$}{
  \texttt{InnerGroupOpt}(group)\;
}
$improved \gets \texttt{true}$\;
\While{$improved$}{
  $improved \gets \texttt{InterGroupOpt}(pathgroups)$\;
}
\ForEach{group $\in pathgroups$}{
  \texttt{InnerGroupOpt}(group)\;
}
$l' \gets \texttt{GetOptimizedLabel}(pathgroups)$\;
\Return $l'$\;
\end{algorithm}

\subsection{Search Process and Post-Optimization}
\label{sec:post-opt}

As shown in Alg.~\ref{alg:saf}, \abbrOurMTSP adds the initial label $l_0$ to the open list $L_{O}$ and focal list $L_{F}$ and begins the search.
In each iteration, the label $l$ with the highest priority in $L_{F}$ is selected for expansion and is removed from both $L_{O}$ and $L_{F}$.
Before expanding $l$, $l$ is checked for dominance against labels in $L(\vec{v}(l))$.
If $l$ is dominated, $l$ is discarded and the current while iteration ends.
Otherwise, $l$ is used to filter $L(\vec{v}(l))$ and then added to $L(\vec{v}(l))$, where filtering means removing any existing labels in $L(\vec{v}(l))$ that are dominated by $l$.
If $l$ visits all nodes and all robots reach their goals $\vec{v}_g$, then a solution is found, which is then post-optimized as explained in the next subsection.
After post-optimization, the inner while loop is broken and $\epsilon$ decreases so as to find a better solution until time out.
Then, $l$ is expanded using the partial expansion as described in Sec.~\ref{sec:global:seq_exp} and a set of successor labels is created.
For each of those successor labels $l'$, $l'$ is checked for dominance.
Additionally, $l'$ is also checked for feasibility in \texttt{NotFeasible} by validating:
\begin{itemize}
    \item (i) whether $l'$ obeys the assignment constraints and
    \item (ii) when all robots reach their goals ($\vec{v}(l)=\vec{v}_g$), if there are still any unvisited nodes.
\end{itemize}
If $l'$ survives these checks, $l'$ is added to $L_{O}$ for future expansion.
The focal list $L_{F}$ is updated correspondingly so that $L_{F}$ contains all labels $l$ in $L_{O}$ such that $f(l)\le(1+\epsilon)\cdot f_{\text{min}}$, with $f_{\text{min}}$ being the minimum $f$-value of any label in $L_{O}$.
\abbrOurMTSP terminates when the runtime limit is reached.

After finding a feasible solution, \abbrOurMTSP runs local optimization (Alg.~\ref{alg:postOpt}) to further reduce the makespan.
Here, the robots are classified into a finite set of types, and robots are grouped by their types.
Specifically, if two robots $i,j\in I, i\neq j$ satisfy $i\in A(v), j\in A(v)$ for all vertices $v\in V$, then robots $i,j$ are of the same type.
For example, two robots are both legged, then they can visit the same subset of vertices in the graph $G$.

Each robot's path is then improved as follows.
\abbrOurMTSP first applies an \emph{inner-group} 2-opt optimization follows to possibly further reduce the makespan.
After that, an additional \emph{inter-group} optimization, where the nodes in the longest path are iteratively selected and transferred to the shortest path in another group with the smallest inner-group makespan. 
Finally, another round of \emph{inner-group} optimization is applied.

\subsection{Analysis}\label{heha:sec:analysis}

\begin{lemma}\label{heha:lem:dom}
    For two labels $l_1,l_2$, if $l_1$ dominates or is equal to $l_2$, $l_2$ cannot lead to a better solution than $l_1$.
\end{lemma}
\begin{proof}

    We prove this lemma by showing that for any solution path $\pi_2$ that extends $l_2$, a corresponding solution path $\pi_1$ that extends $l_1$ can be constructed and is no worse than $\pi_2$.
    Let $\pi_a(l_1),\pi_a(l_2)$ denote the joint paths of all robots from $\vec{v}_s$ to $\vec{v}(l_1)$ and $\vec{v}(l_2)$, respectively.
    Let $\pi_b(l_2)$ denote an arbitrary joint path starting from $\vec{v}(l_2)$ that visits all unvisited vertices in $V\setminus B(l_2)$.
    Thus, the concatenated joint path $\pi_2=\pi_a(l_2)+\pi_b(l_2)$ is a solution extending $l_2$.
    Since $\vec{v}(l_1)=\vec{v}(l_2)$ (i.e., each robot is at the same vertex in $l_1$ as in $l_2$), joint paths $\pi_a(l_1)$ and $\pi_b(l_2)$ can be concatenated.
    Let $\pi_{12}=\pi_a(l_1)+\pi_b(l_2)$ denote the resulting joint path.
    Since the accumulated cost of each robot in $l_1$ is no greater than that of the corresponding robot in $l_2$, and $\pi_{12}$ and $\pi_2$ share the same continuation $\pi_b(l_2)$, we have $g_{\text{max}}(\pi_{12})\leq g_{\text{max}}(\pi_2)$.
    However, since $B(l_1)\supseteq B(l_2)$, some vertices visited in $\pi_b(l_2)$ may have already been visited in $\pi_a(l_1)$, and therefore may be visited more than once in $\pi_{12}$.

    For each vertex $u_k$ that is visited more than once in $\pi_{12}$, if $u_k$ is the last vertex of some robot $i$'s path $\pi^i$, then this occurrence of $u_k$ can be removed without increasing $g_{\text{max}}(\pi_{12})$, since $\pi^i$ is shortened while all other robot paths remain unchanged.
    Otherwise, let $u_{k-1}$ and $u_{k+1}$ denote the previous and next vertices of $u_k$ in robot $i$'s path $\pi^i$.
    The occurrence of $u_k$ can then be shortcut by directly connecting $u_{k-1}$ and $u_{k+1}$ and skipping $u_k$.
    By the triangle inequality, this shortcut does not increase the length of $\pi^i$, while all other robot paths remain unchanged.
    Therefore, the resulting joint path has a $g_{\text{max}}$ value no greater than that before the shortcut.
    We repeatedly apply this operation until all redundant visits in $\pi_{12}$ are removed.
    Let $\pi_1$ denote the resulting joint path after all such shortcuts.
    Since only redundant visits are removed, $\pi_1$ still visits all required vertices and is therefore a solution extending $l_1$.
    Moreover, $g_{\text{max}}(\pi_1)\leq g_{\text{max}}(\pi_{12})\leq g_{\text{max}}(\pi_2)$.
    Thus, for any solution extending $l_2$, there exists a solution extending $l_1$ that is no worse.
    $l_2$ thus cannot lead to a better solution than $l_1$ and is pruned.
\end{proof}

\begin{theorem}
    \abbrOurMTSP is complete in a sense that, (i) for an unsolvable mHPP instance, \abbrOurMTSP terminates in finite time and reports failure; (ii) for a solvable mHPP instance, \abbrOurMTSP returns a feasible solution in finite time. 
\end{theorem}

\begin{proof}
\abbrOurMTSP conducts systematic best-first search as A*~\cite{A*_study}, where in each search iteration the most promising label in FOCAL is selected for expansion, and when expanding a state, \emph{all} possible successor states are considered.
During the search, a label $l$ is pruned only when there is another label $l'$ that dominates $l$.
By the definition of dominance (Def.~\ref{heha:def:dom}) and Lemma~\ref{heha:lem:dom}, such pruning only removes sub-optimal solutions since for any future solution from $l$, there must exist a corresponding (better) solution from $l'$.
As a result, for a solvable problem instance, \abbrOurMTSP never eliminates the only solution joint path if a better path has not been found yet, and terminates the search with a feasible solution.
For an unsolvable problem instance, since the graph is finite, there is a finite number of possible joint paths.
\abbrOurMTSP terminates after enumerating all joint paths and reports failure.
\end{proof}

\begin{theorem}
    Let $g_*$ denote the cost of the true optimal solution, \abbrOurMTSP finds a $(1+\epsilon)$ bounded sub-optimal solution $\pi$ for any solvable Min-Max mHPP as defined in Problem~\ref{heha:problem:mhpp}, i.e., $g_{\text{max}}(\pi) \leq (1+\epsilon)g_*$.
\end{theorem}
\begin{proof}
    \abbrOurMTSP systematically enumerates all paths in the joint graph $\mathcal{G}$, and by Lemma~\ref{heha:lem:dom}, dominance pruning never eliminates a path along an optimal solution.
    With an admissible heuristic (Lemma~\ref{heha:lem:heu_adm}), the $f$-value of any label $l$ in OPEN is a lower bound of the true optimal solution cost~\cite{A*_study}, i.e., $f(l)\leq g_*$ with $g_*$ being the true optimal cost of an optimal solution.
    The use of focal search~\cite{A*_study} allows \abbrOurMTSP to select a label $l$ with $f$-value within range $[f_{\text{min}},(1+\epsilon)f_{\text{min}}]$ with $f_{\text{min}}$ being the minimum $f$-value of any label in OPEN.
    At termination, the popped label $l$ is guaranteed to have $g_{\text{max}}(l) = f(l) \leq (1+\epsilon) f_{\text{min}} \leq  (1+\epsilon) g_*$, which ensures the $(1+\epsilon)$ bounded sub-optimality of the returned solution.
    The use of \texttt{PostOptimize} only further improves the solution quality of the popped label and will never increase the cost.
\end{proof}

\section{Other Components in \abbrName}
\label{sec:other_components}

\subsection{Terrain Traversability}
\label{sec:terrain}

\subsubsection{Terrain Analysis}
\begin{figure}[tb]
    \centering
    \includegraphics[width=0.88\linewidth]{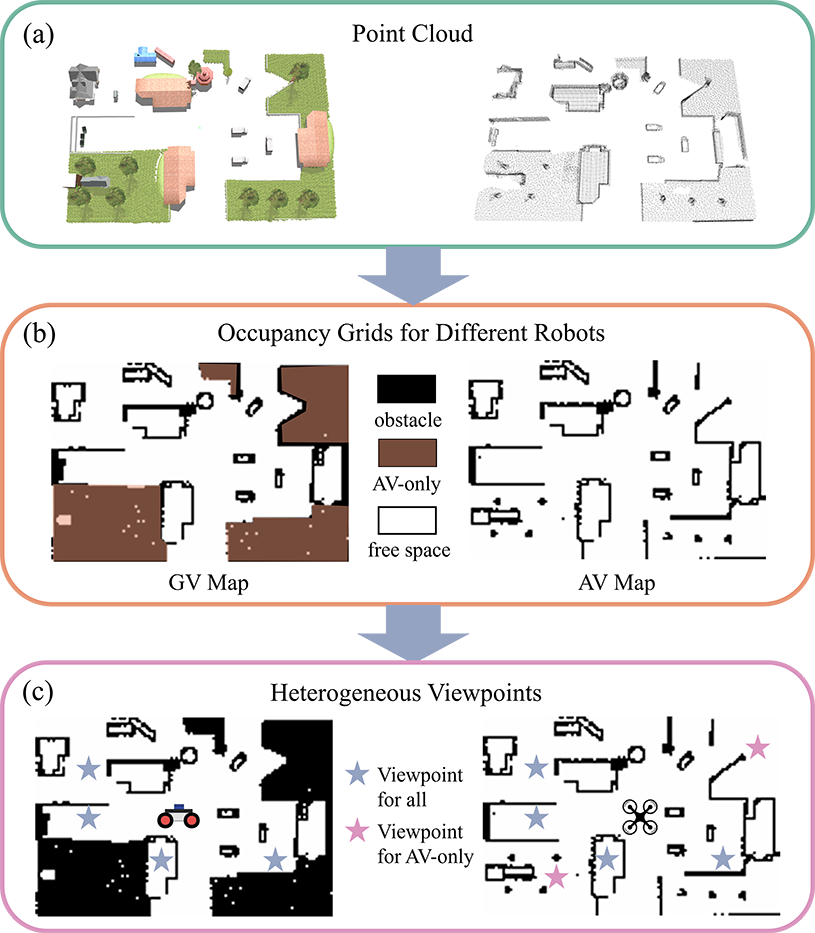}
    \caption{A visualization of the terrain analysis process in \abbrName. Multiple occupancy grids are generated for different types of robots, and viewpoints are sampled for different robots with different traversability, which are then used for the global planning.}
    \label{fig:terrain}
\end{figure}

In \abbrName, due to the heterogeneous terrain traversability of different robot platforms, different occupancy grids are maintained for different robot types, as shown in Fig.~\ref{fig:terrain}.
For ground robots, including wheeled and legged robots, these grids encode terrain-dependent traversability, while for UAVs, the corresponding grid mainly reflects the availability of free flying space based on obstacle and height information.
All these occupancy grids share the same spatial information about known and unknown cells at any time during exploration.
To build the occupancy grids for ground robots, we use a plane-fitting method~\cite{putn}.
To summarize the main ideas, point cloud data is first acquired through SLAM software and uniformly sampled with a fixed spatial interval.
For each sampled block, singular value decomposition (SVD) is performed to obtain the dominant direction vectors that describe the local geometric structure of the point cloud.
By evaluating the consistency of these normal vectors, points corresponding to the surfaces of the ground or vertical walls are filtered out.
Subsequently, the remaining point cloud is analyzed based on its normal distribution to extract regions exhibiting significant inclination or step-like geometry, such as slopes and staircases.

The identified regions are then projected onto the respective grid maps of each type of robot. 
For the wheeled robot, overly steep or uneven areas are marked as occupied cells, representing non-traversable regions for the wheeled robots. 
In contrast, the same areas are labeled as traversable in the legged robot's grid, since the legged platform is capable of passing through them.
We call these areas \emph{heterogeneous regions}, which are areas that are traversable for the legged robot but not for the wheeled robots. 
Viewpoint sampling is then performed within these heterogeneous regions to generate \emph{heterogeneous viewpoints}, which are poses that are only accessible for certain types of robots. 
For example, some of those heterogeneous viewpoints can be explored by the legged robot and are inaccessible to the wheeled robots.
\subsubsection{Failure Handling}

During exploration, if no terrain analysis result is received by the planner, then there is no way for \abbrName to determine which type of robot can traverse that region.
Therefore, the region is treated as an obstacle (i.e., non-traversable region) for all robots.
In this region, no free space will be created in the resulting map, and no viewpoints will be generated in that region.

If a non-traversable region is mistakenly classified as traversable, the robot will usually stop its motion as it approaches that non-traversable region due to the local obstacle collision avoidance mechanism, and be able to find no feasible path exists in that region in the subsequent detection-planning cycle. In that case, the viewpoints in that non-traversable region are discarded.
In the worst case, if the local collision avoidance fails, then it indicates that there is no way for the robot to identify that non-traversable region and the robot will have to collide or get trapped.

\subsection{Local Planning}
\label{sec:local}

\begin{figure}[tb]
    \centering
    \includegraphics[width=0.88\linewidth]{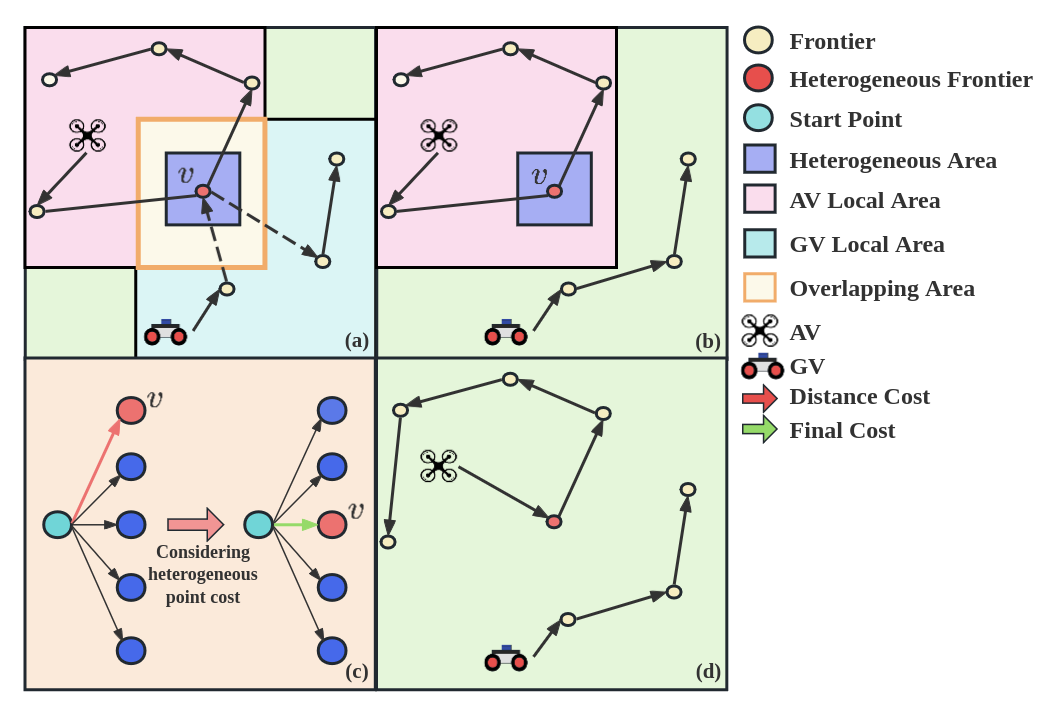} 
    \caption{An illustration of two tricks in Local Planning. (a) shows the set of local viewpoints and the baseline path generated without specialized planning constraints; a specific node $v$ in heterogeneous region (painted as violet) is identified within the overlapping local areas of both robots. (b) shows that $v$ is assigned to the AV due to the high priority of AV. (c) details the initial phase of local planning, where the TSP cost matrix is adjusted to incorporate hetero-frontier cost, encouraging an optimal approach to the viewpoints. (d) displays the final optimized local plan resulting from the cost-adjusted TSP.}
    \label{fig:p2}
\end{figure}

Let $D^i \subseteq \mathcal{W}$ denote a bounded space around the robot $i\in I$ for local planning.
During exploration, there are often many frontiers near each other; we therefore down-sample these frontiers.
Specifically, for a set of frontiers $Q(D^i)$ to be visited by robot $i$ (i.e., $i\in A(q), q\in Q$), \abbrName clusters them into subsets $Q(D^i)=Q_1\cup Q_2\cup \cdots \cup Q_K$ based on the position of each frontier $q\in Q(D^i)$, so that the frontiers in each $Q_k, k=1,2,\cdots,K$ are close to each other.
Within each $Q_k$, a representative frontier $q_k'$ near all other frontiers in $Q_k$ is selected for the robot to visit.
Then, similar to \cite{tare}, a TSP is formulated and solved to obtain a local path $\pi^i_{\text{local}}$ for robot $i$ to visit all these representative frontiers.

Specifically, let $G_L=(V_L,E_L)$, $L$ stands for local, denote a complete graph where $V_L$ is a set of vertices, one for each representative frontier $q_k', k=1,2,\cdots,K$, and $E_L=V_L\times V_L$ is a set of edges that represent the transition from one vertex to another.
In $G_L$, each edge $e \in E_L$ is associated with a non-negative cost value $c(e)$ representing the transition path cost from one representative frontier to another, which is estimated by using Hybrid-state A*~\cite{dolgov2010path} to find a kinodynamically feasible trajectory from one frontier to another.
Then, a TSP problem can be defined on $G_L$ which seeks a cycle in $G_L$ that visits all vertices and returns to the starting vertex.

When exploring the frontiers in $D^i$, new frontiers may occur and the robot iteratively replans path locally.
Thus, the robot only executes the sub-path to reach the first frontier in $\pi^i_{local}$ and then a new TSP is formulated and solved, which includes the new frontier if any. 

Besides these similar components to~\cite{tare}, \abbrName has two additional tricks to handle the heterogeneous robots in the local planning as detailed next.

\subsubsection{Hetero-Frontier Cost}
In the graph $G_L$ for local planning, each vertex in $G_{L}$ corresponds to either the current pose of the robot (denoted as $v_0$), or a representative frontier $q'_k$ as aforementioned.
Some of the edge costs are adjusted to encourage early exploration of frontiers that can only be visited by a subset of agents due to the terrain and traversability of the robots.
A frontier $q$ is a \emph{hetero-frontier} for a robot $i\in I$, if $A(q) \neq I$ and $i\in A(q)$, i.e., $q$ can only be visited by a subset of robots and the current robot $i$ being planned can visit it.
When down-sampling the frontiers in each subset $Q_k$ as aforementioned, \abbrName also counts the percentage $\alpha\in[0,1]$ of hetero-frontiers in $Q_k$.
For each edge $e$ from $v_0$ to some other vertex $v\in G_{L}$, its edge cost is changed to 
\begin{align}
    c'(e):= \max\{c(e)-\alpha \frac{n_h}{n_0},0\}\label{heha:eq:hetero_cost}
\end{align}
where $c(e)$ is the original edge cost of the edge $e=(v_0,v)$, $n_h$ is the number of hetero-frontiers within a certain range (e.g. a circle of radius $r_h \geq 0$ and $r_h=2$ in our tests) from $v$, and $n_0$ is a user-defined non-negative number, which is set to $25$ in our tests.
By doing so, the edges to the hetero-frontier tend to have lower cost, which thus encourages the robot to visit them as the next step in a soft manner when solving the TSP problem.
Here, Eq.~(\ref{heha:eq:hetero_cost}) is just an empirical way to realize the idea of encouraging early exploration of hetero-frontier and other expressions are also possible.

\subsubsection{Priority Assignment}
When two robots' local planning spaces intersect (i.e., $D^{i,j}:=D^i\cap D^j \neq \emptyset$) and there are hetero-frontiers in $D^{i,j}$ (i.e., $\exists q\in Q(D^{i,j})$), then the local planner needs to determine which robot should visit those hetero-frontiers.
\abbrName introduces additional priorities $\xi^i,i\in I$ to all robots, and those hetero-frontiers are assigned to the robots with higher priority.
In practice, for example, we set the priority of drones to be higher than legged and legged to be higher than wheeled, since drones can traverse all terrains while legged and wheeled can only access a subset of terrains.
These priorities help reduce redundant and duplicated exploration by different robots in overlapping local regions and therefore improve the exploration efficiency.

\subsection{Planning Cycles in \abbrName}
\label{sec:local_global_interaction}

The global and local planners in \abbrName operate as follows.
First, the local planner is invoked with a fixed rate (once per 2.5 seconds, $0.4-0.5\text{ Hz}$).
Each invocation of the local planner forms a $L\times L$ (e.g. $L=7 m$) local squared region centered at the current position of the robot and plans a local path for the robot to visit all frontier viewpoints in this region. In addition to this periodic invocation, HEHA also supports condition-triggered replanning through a 2 Hz checking mechanism. For example, if frontier viewpoints in the local region disappear rapidly due to exploration progress or map updates, replanning can be triggered earlier without waiting for the next fixed-period cycle.

Second, when the number of frontier viewpoints in the local square region is below a certain threshold (e.g. when the robots run into a dead-end or when surrounding areas are already explored), or when the local planner fails to plan a path (e.g. due to non-traversable terrains), the global planner is invoked, which formulates a mHPP problem and uses the proposed \abbrOurMTSP to quickly solve it.
The resulting path can provide the robot with global guidance.

With such designs, in practice, we observed that, at the beginning of the exploration, the local planner is often able to quickly generate paths while locally coordinating with other robots of the same or different types to avoid duplicated exploration of overlapped areas.
As the exploration progresses into the middle and late stages, the distribution of frontiers becomes increasingly sparse and distant.
Consequently, the local planner fails more frequently, which calls the global planner more frequently.
Sometimes, the frequency of calling the global planner gradually increases until it aligns with the local planning frequency (approximately $0.4-0.5\text{ Hz}$).

In each global planning, an mHPP is formulated based on the current map and frontiers, and the solution to this mHPP is computed from scratch.
The fast running speed of our \abbrOurMTSP makes such online ``real-time'' global planning on robots possible.

\section{Experimental Results}

We compare our global planner \abbrOurMTSP against baselines for mHPP, compare our \abbrName against baselines in simulations for heterogeneous multi-robot exploration, and test the robustness and sensitivity of \abbrName in simulation.
Then, we conduct an ablation study to verify the benefits of the two tricks for the local planning, and finally demonstrate our \abbrName with real robots as shown in the video attachment.

\subsection{Global Planning and mHPP}
\label{src:h-mTSP_test}

\begin{figure}[tb]
    \centering
    \includegraphics[width=0.88\linewidth]{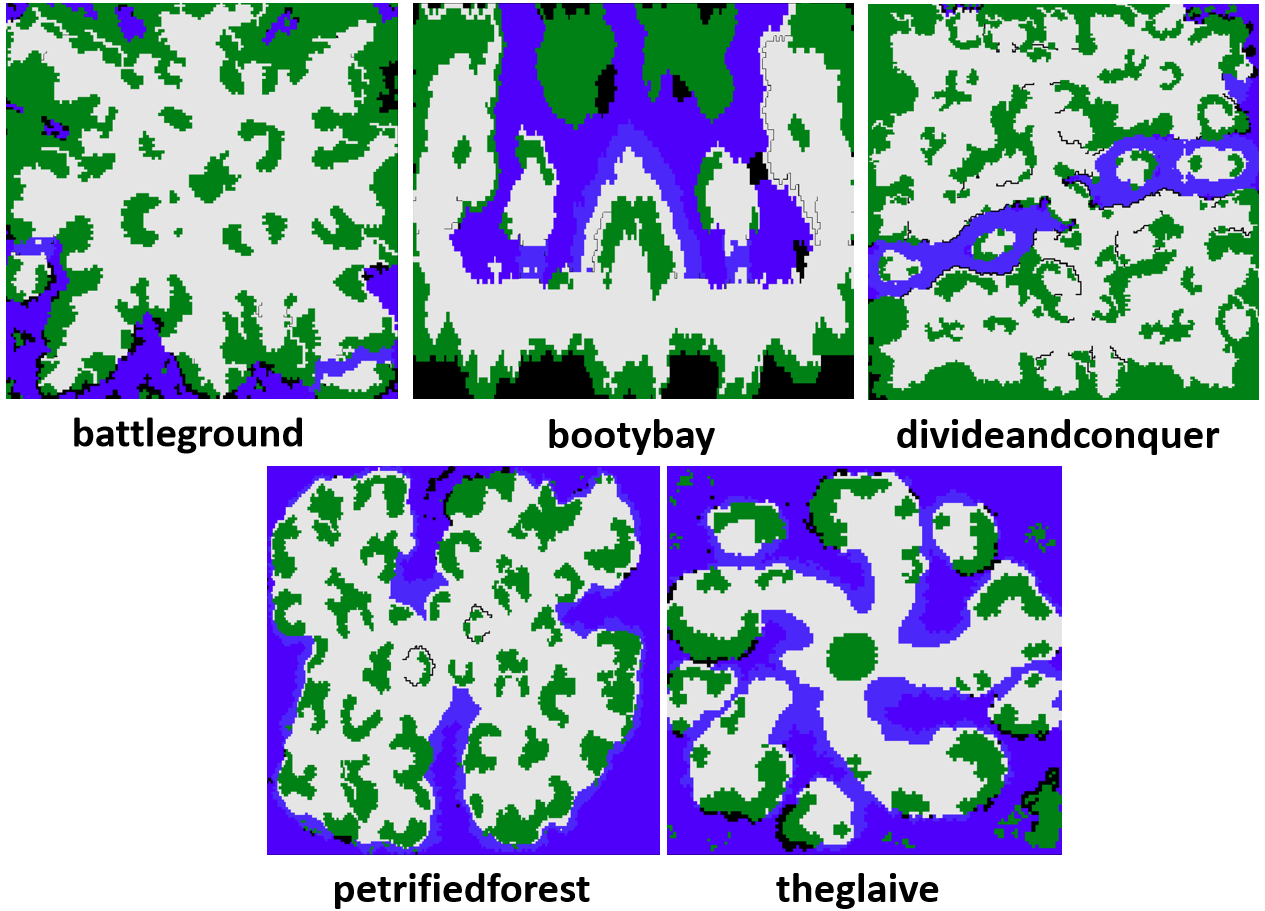}
    \caption{Five grid maps of size $512\times 512$ from a public dataset, which is used in our experiments to generate mHPP instances.}
    \label{fig:maps}
\end{figure}

\begin{figure*}[t]
    \centering
    \includegraphics[width=0.88\linewidth]{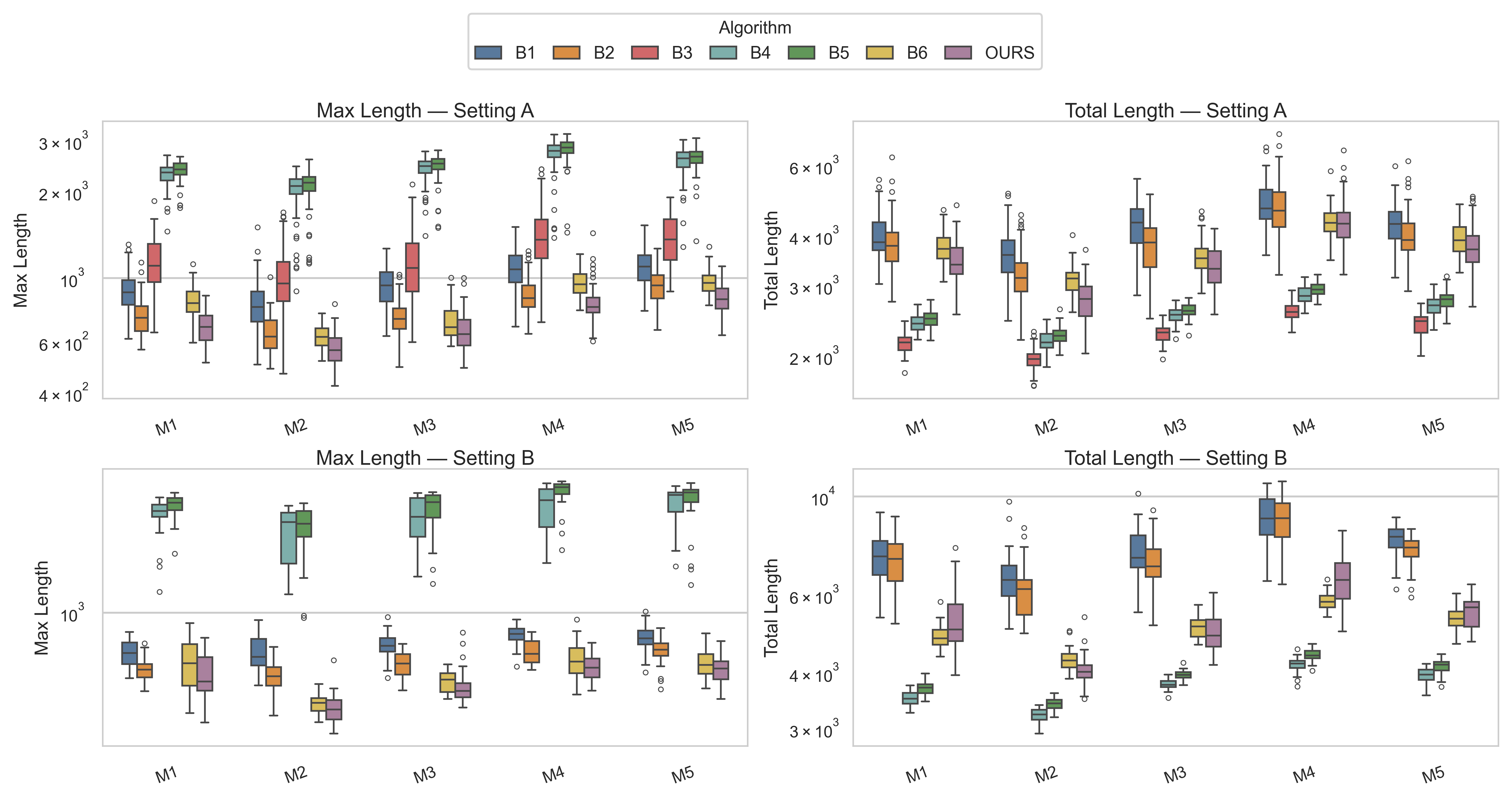}
    \caption{Makespan (i.e., maximum length of all robots' path) and Sum of costs (i.e., total length of all robots' path) comparison under Setting A and B in five maps.
    Colors of algorithms (B1–B6 and OURS) are consistent across both settings.}
    \label{fig:box_max_sum}
\end{figure*}

\subsubsection{Maps and Test Settings}
We use five grid maps (Fig.~\ref{fig:maps}) of size $512 \times 512$ cells from a public dataset \cite{sturtevant2012benchmarks}, where each map is a 2D grid with heterogeneous terrain types, including ground, forest, swamp, water, and obstacle. For simplicity, we denote the five grid maps as M1–M5 in the same order as shown in Fig.~\ref{fig:maps}.
We consider two types of robots, ground vehicle (GV) and aerial vehicle (AV).
To generate an instance, we sample a few nodes from these grid maps to be visited by the robots, and generate a complete graph, where the edge costs are the shortest path cost between these sampled nodes using A* search in the grid map.
These sampled nodes may be located on different terrains, and thus lead to assignment constraints.
We classify the nodes to be visited by the robots into two classes: (i) \textit{Normal nodes}, which are sampled from \textit{ground} grids so that both GVs and AVs can visit, (ii) \textit{AV-only nodes}, which are sampled from \textit{swamp} and \textit{water} grids that can only be visited by AVs.
The ratio of normal nodes and AV-only nodes is 2:1.

The experiments involve two different settings: \textit{Setting A} has 60 nodes, 3 GVs and 3 AVs;
\textit{Setting B} has 150 nodes, 10 GVs and 10 AVs.
Both Setting A and B are repeated 30 times, with nodes randomly sampled each time.
In our tests, we set a runtime limit of 30 seconds for each planner for each instance.

\subsubsection{Baselines}
We use six baselines (B1-B6).
\begin{itemize}
    \item 
B1 is a greedy method that incrementally assigns nodes to the robots subject to assignment constraints while minimizing the current makespan.
    \item 
B2 builds upon B1 and further uses the \texttt{PostOptimize} (Alg.~\ref{alg:postOpt}) to refine the solution returned by B1.
The comparison between B2 and B1 will show the improvement on solution quality provided by the local optimization idea in the proposed \texttt{PostOptimize}.
    \item 
B3 uses the transformation method in \cite{today_TSP} to convert mHPP to a single-agent TSP on a larger graph $G'$ and then invokes LKH~\cite{k_opt_lkh}, a popular solver for routing problems, to find a solution that minimizes the sum of costs for mHPP.
In this transformation method, the transformed graph $G'$ depends on both the number of agents and the size of the graph.
In our experiments, this method is only able to handle Setting A.
For Setting B the graph $G'$ is too big for this method to return any solution within the runtime limit, which is thus omitted from the results as shown next.
    \item 
B4 uses another popular solver PyVRP \cite{pyvrp} to solve mHPP with assignment constraints while minimizing the sum of costs.
In our tests, we set the termination condition of B4 as follows: B4 terminates after 1000 iterations without additional improvement on the incumbent solution.
    \item
B5 uses Google OR-Tools~\cite{ortools_routing} with default parameters, which optimizes the sum of costs.
    \item
B6 is a fine-tuned version of B5, including an augmented objective that encourages balanced workloads and a smaller makespan.
Specifically, it adds a penalty term proportional to the difference between the longest and shortest path lengths:
$\text{Objective}=\sum_{i\in I} L_i + 100 \cdot (\max_{i\in I} L_i - \min_{i\in I} L_i)$,
where $L_i$ denotes the path length of robot $i$.
This penalty encourages more balanced workloads among robots and leads to smaller makespan.
\end{itemize}

 The six baselines above can be grouped into three categories according to their optimization objectives and roles in the evaluation.
First, B1 and B2 optimize makespan, sharing the same min-max objective as \abbrOurMTSP.
Second, B3--B5 retain the native min-sum objectives of representative and widely used routing solvers. 
These baselines are included to assess the consequence of applying existing high-performance min-sum solvers to our problem, where the makespan is the focus. 
Note that, B3--B5 are not intended as strictly same-objective comparisons with our approach, but serve as practical references for solution quality and computational efficiency.
Third, B6 represents adaptation of an existing min-sum solver to the min-max objective. 
Its comparison with \abbrOurMTSP evaluates whether such an adaptation can achieve performance comparable to a solver specifically designed for constrained min-max mHPP.

\subsubsection{Solution Makespan Comparison}
For the experimental results, among the instances, we report their average total cost (avg sum), average makespan (avg max), and average search time (avg time), i.e., the last time point when an algorithm updates the solution within the runtime limit.

As shown in Fig.~\ref{fig:box_max_sum}, our \abbrOurMTSP always finds solutions with better makespan on average (avg max).
Detailed distributions of solution makespans for both settings are shown with box plots in Fig.~\ref{fig:box_max_sum}.
These results verify that \abbrOurMTSP is able to find solutions with a better makespan than the existing methods, which, as shown later, helps improve the completion time of exploration in various environments.

\subsubsection{Solution Sum of Costs Comparison}
\begin{figure}[t]
    \centering
    \includegraphics[width=0.88\linewidth]{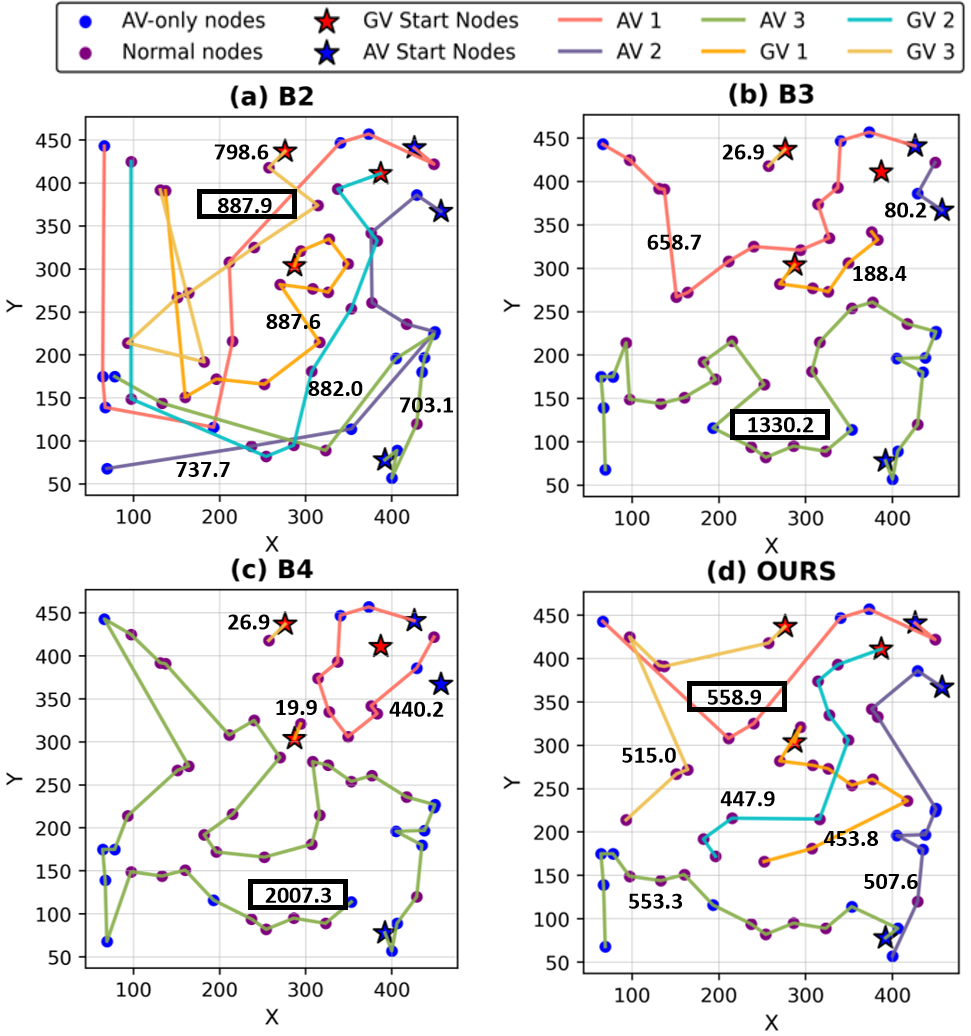}
    \caption{Visualization the solution paths of baseline B2, B3, B4 and OURS on an instance with Setting A. The boxed number is the makespan. Our \abbrOurMTSP can distribute the nodes in a more balanced fashion among the robots and thus achieves smaller makespan.}
    \label{fig:vis_algo}
\end{figure}

We also report the sum of costs (i.e., total length) returned by different planners in Fig.~\ref{fig:box_max_sum}.
It is not a surprise that the existing planners B3 (LKH transformation) and B4 (PyVRP) find solutions with much smaller sum of costs than our \abbrOurMTSP, since they are designed to optimize sum of costs.
We visualize the solutions for an instance in Fig.~\ref{fig:vis_algo}.
We observe that the solutions with small sum of costs returned by B3 and B4 can lead to very unbalanced path lengths among the robots, resulting in a large makespan.
When applied to exploration, these unbalanced solutions can increase the completion time of exploration.

\subsubsection{Effects of \texttt{PostOptimize}}
We observe that B2 finds a better solution than B1, due to the local optimization, which verifies the benefits of applying \texttt{PostOptimize}.
Our \abbrOurMTSP is better than B2, since \abbrOurMTSP conducts systematic search while employing \texttt{PostOptimize}, and B2 can get stuck in local minima that is near the initial greedy solution.

\begin{figure}[t]
    \centering
    \includegraphics[width=\linewidth]{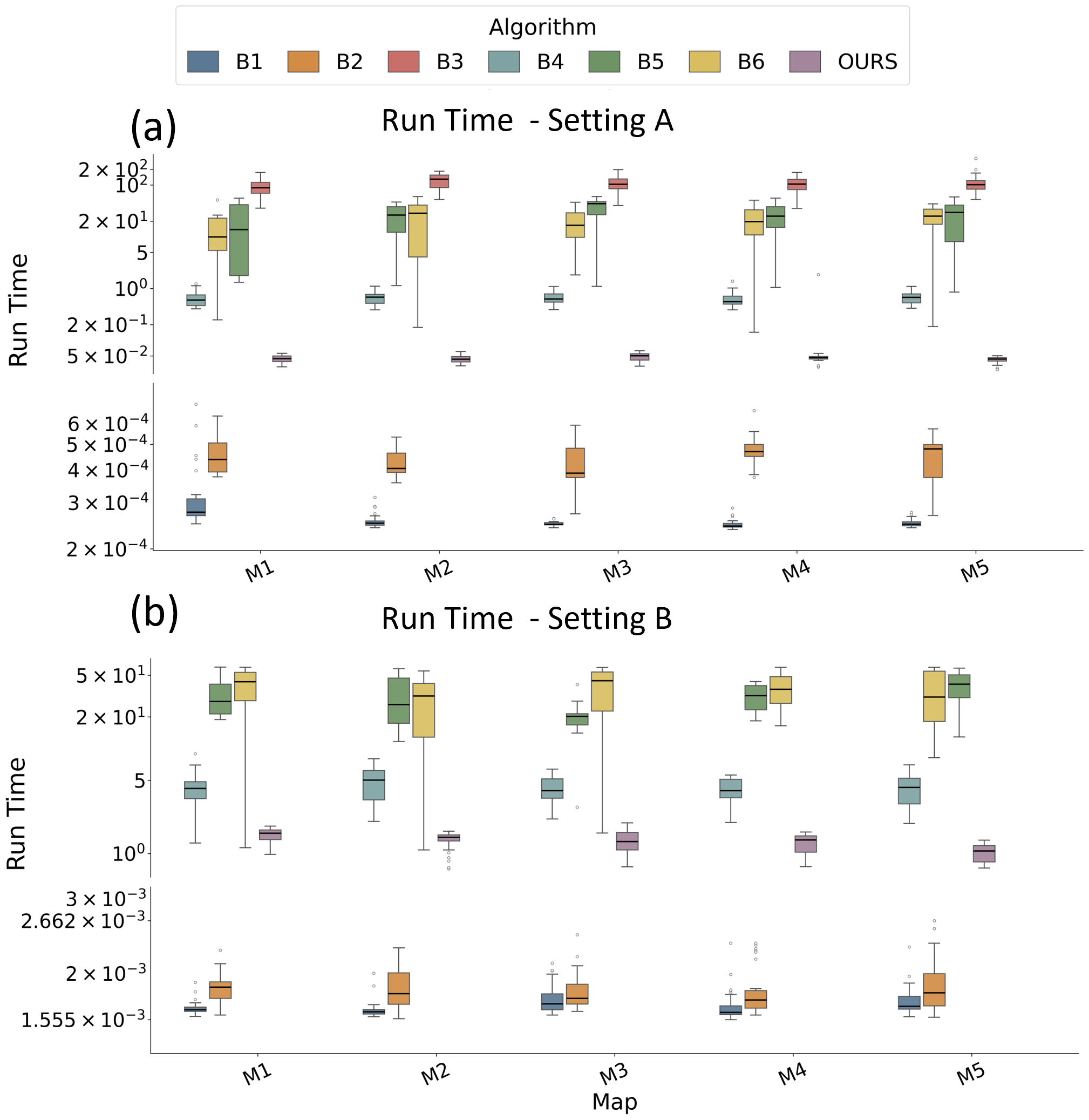}
    \caption{Time to get final solution under Setting A and B in five maps.
    Colors of algorithms (B1–B6 and OURS) are consistent across both settings.}
    \label{fig:time}
\end{figure}

\subsubsection{Runtime Comparison}
As shown in Figure~\ref{fig:time}, B1 and B2 runs the fastest due to their simple greedy strategy.
Our \abbrOurMTSP is slower than the greedy methods B1 and B2, but is often obviously faster than B4, B5 and B6.
Recall that the reported runtime of \abbrOurMTSP here is the last time point when an algorithm updates the solution within the runtime limit.
It indicates that, in practice, \abbrOurMTSP is able to find high-quality solutions very quickly.
 Note that, the runtime reported in Figure~\ref{fig:time} is the computational time of the routing solver, rather than the exploration completion time. 
This metric is relevant because the global mHPP is repeatedly solved online during exploration, requiring the planner to return high-quality routing solutions quickly. 
The practical exploration completion time is evaluated separately in Sec~\ref{sec:sim_exp}.

\begin{figure}[t]
    \centering
    \includegraphics[width=0.88\linewidth]{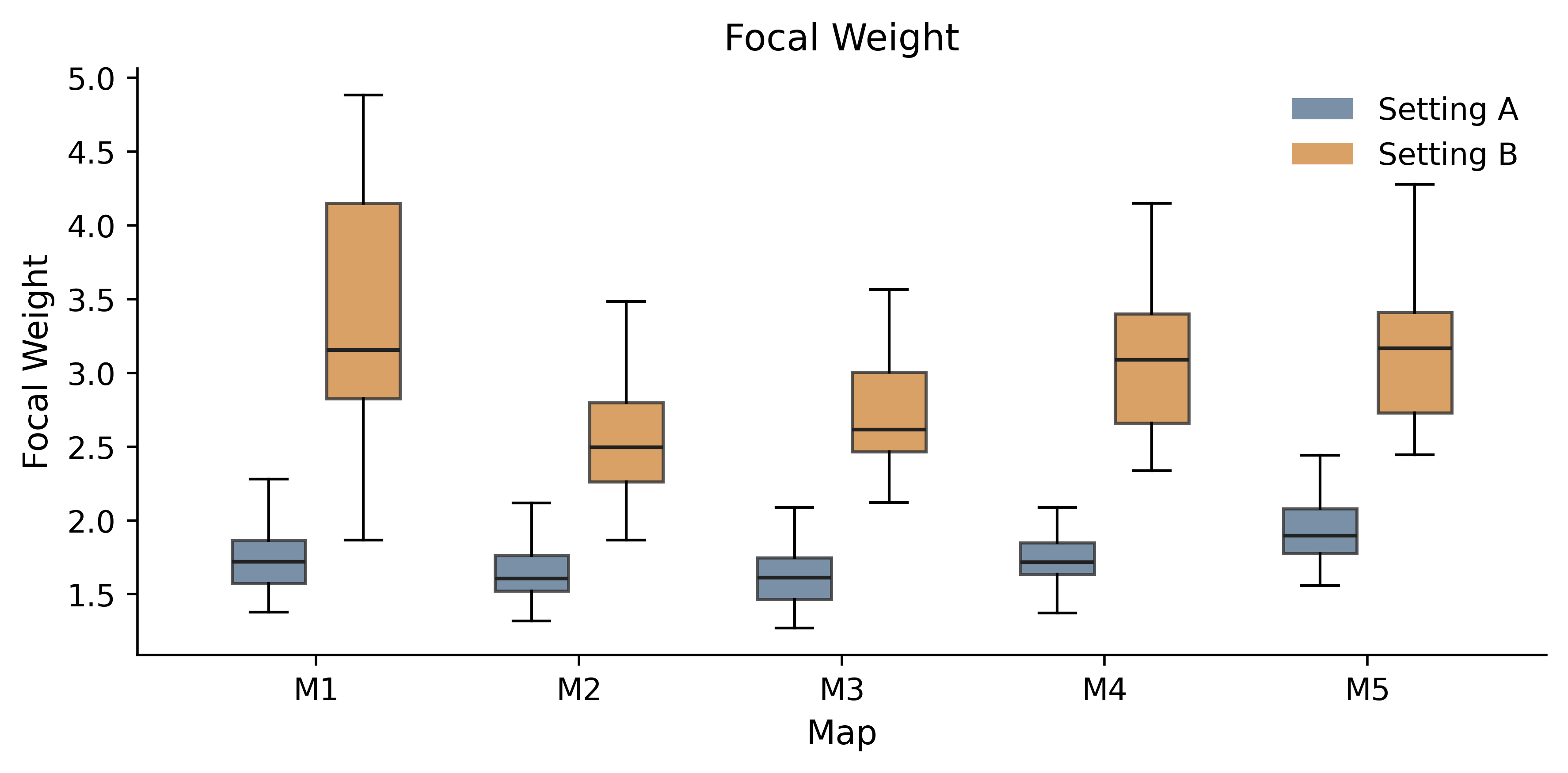}
    \caption{Focal weight $\epsilon$ when \abbrOurMTSP terminates under Setting A and B in five maps, which provides a sub-optimality bound of the returned solutions.}
    \label{fig:focal_ratio}
\end{figure}

\subsubsection{Solution Sub-Optimality Bounds}

Since \abbrOurMTSP is an anytime algorithm where the focal weight $\epsilon$ decreases iteratively, we set the initial focal weight $\epsilon$ to a large number and report the ending $\epsilon$ when \abbrOurMTSP terminates in Fig.~\ref{fig:focal_ratio}.
Note that this weight $\epsilon$ bounds the solution sub-optimality. 
The relatively large values observed are mainly due to the scale of the problem instances in Setting B, where the number of nodes is large, while in Setting A, the bound is relatively tighter.

\subsubsection{Scalability}
\begin{figure}[t]
    \centering
    \includegraphics[width=0.88\linewidth]{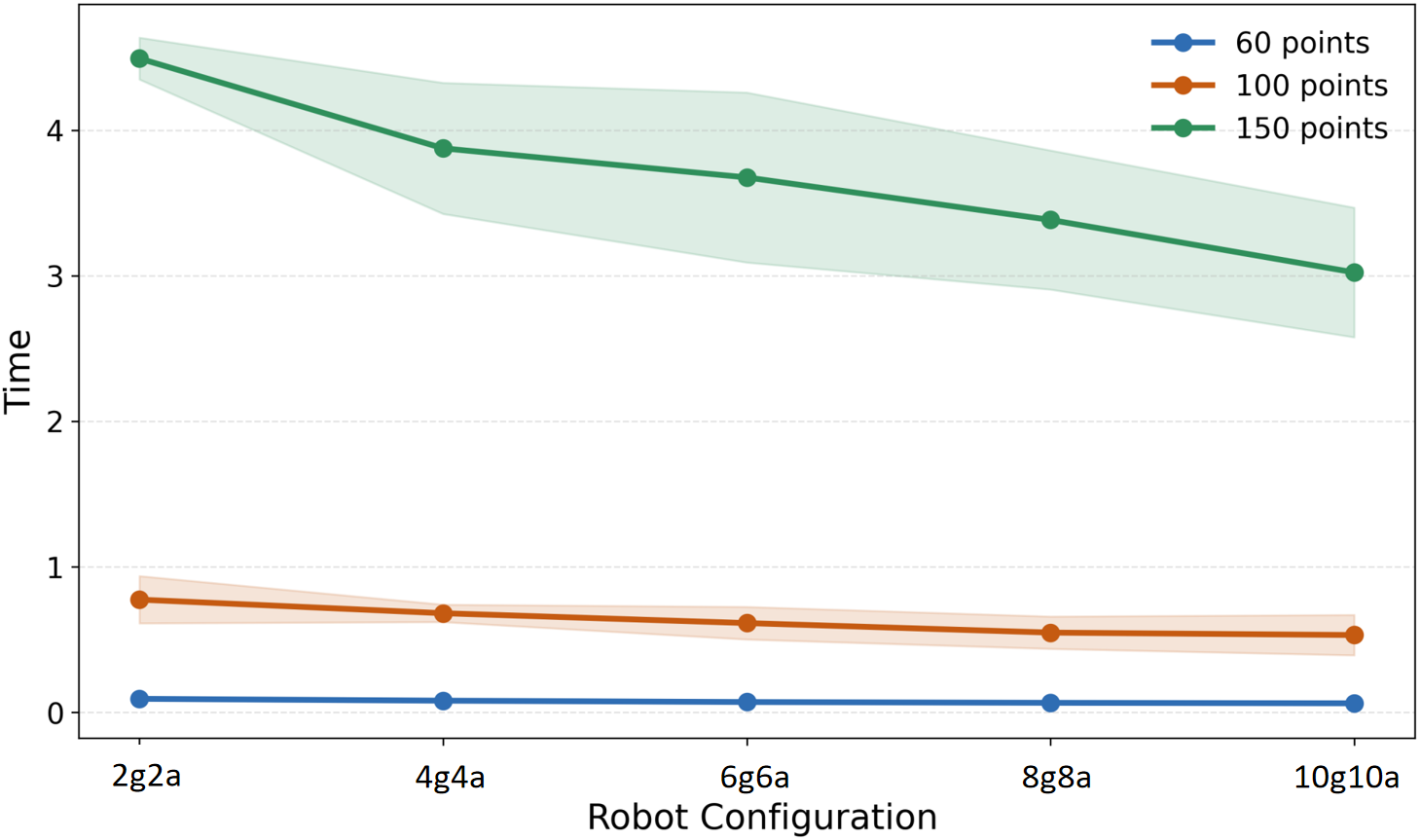}
    \caption{Average solving time on Battleground under varying problem sizes and robot configurations.}
    \label{fig:scale}
\end{figure}

To evaluate the scalability of our method, we fix the map as M1 Battleground and vary the number of robots from 2 GVs and 2 AVs (2g2a) to 10 GVs and 10 AVs (10g10a).
As shown in Fig.~\ref{fig:scale}, the solving time generally increases with the number of target nodes. However, for a fixed number of nodes, the solving time does not increase (and sometimes decreases slightly) as the number of robots increases.
The possible reason for this trend is due to the design of our algorithm: at each step, the search expands the robot with the currently minimum cost.
As a result, increasing the number of robots allows the path of one robot to finish early, which then reduces the number of branches and iterations for the future search, resulting in fewer search iterations in total and reducing the solving time.

\subsection{Simulation Experiments on Exploration}
\label{src:simulation_experiment}

\subsubsection{ Comparison with Baselines}
\label{sec:sim_exp}

\begin{table}[tb]
\caption{Results of exploration in simulation in three maps shown in Fig.~\ref{fig:sim_res}.
The last column Margin shows the percentage of reduction in exploration time and path length of our \abbrName over the best results among the baselines (Nearest, NBVP).}
\centering
\begin{tabular}{l c c c c}
\toprule
\textbf{Map and Metric} & \textbf{Nearest} & \textbf{NBVP} & \textbf{Ours} & \textbf{Margin}\\
\midrule
\textbf{Garden (51m $\times$ 51m)} & & & \\
Time Avg (s) & 255.2 & 284.0 & \textbf{249.4} & 2.3\% \\
Length Avg (m) & 480.5 & 494.1 & \textbf{309.1} & 35.7\% \\
\midrule
\textbf{Village (81m $\times$ 121m)} & & & \\
Time Avg (s) & 1189.4 & 1024.6 & \textbf{715.0} & 30.2\% \\
Length Avg (m) & 2095.8 & 1766.4 & \textbf{1204.5} & 31.8\% \\
\midrule
\textbf{Forest (101m $\times$ 101m)} & & & \\
Time Avg (s) & 880.8 & 652.4 & \textbf{572.8} & 12.2\% \\
Length Avg (m) & 1733.5 & 1276.6 & \textbf{1089.3} & 14.7\%\\
\bottomrule
\end{tabular}
\label{tab:map_comparison}
\end{table}

\begin{figure*}[tb]
    \centering
    \includegraphics[width=\linewidth]{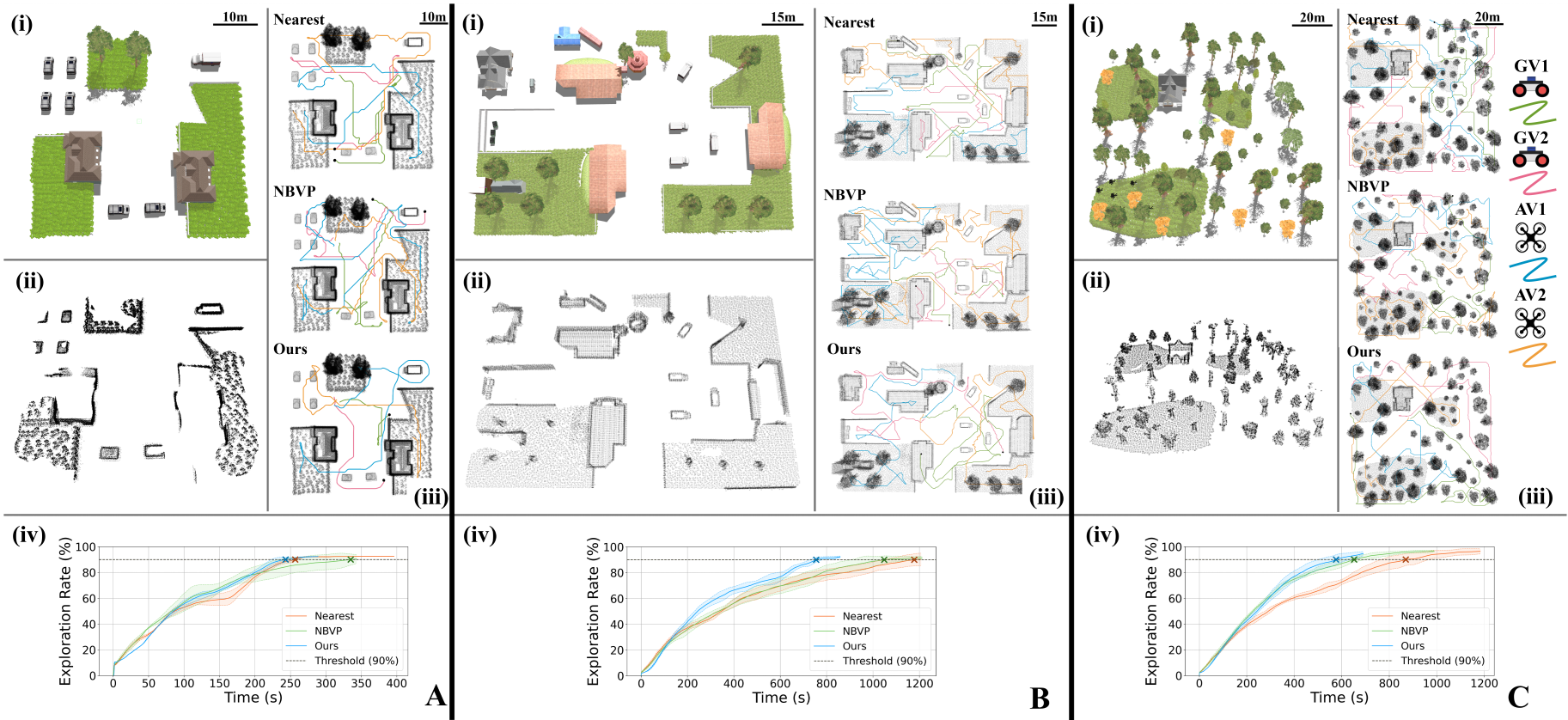} 
    \caption{Exploration in three environments: Garden (A), Village (B), Forest (C). (i) shows the overview of simulation environment. (ii) shows the point cloud map after exploration. (iii) shows the trajectories of the robots using \textit{Nearest}, \textit{NBVP} and Ours. (iv) shows the exploration rates over time.}
    \label{fig:sim_res}
\end{figure*}

We compare our \abbrName against two baselines for exploration.
The first baseline \textit{Nearest} adapts the approach in \cite{frontier_based} by greedily assigning a robot to the closest frontier it can visit.
The second baseline \textit{NBVP} extends \cite{nbvp} to multi-robot, where a Rapidly-exploring Random Tree is built in the free space for each robot with view pose selected to maximize the information gain.
For our \abbrName, we set $\alpha=0.6$ for large maps and $\alpha=0.4$ for small map.
We use the same path planning and trajectory tracking modules across all methods.
All methods terminate after the exploration rate is over 90\% in a similar way to~\cite{tare}, where the exploration rate is the ratio of explored volume over the total volume of the space.
In simulation, we have access to the ground truth environment and can thus compute the volume.
We stop the exploration with a 90\% exploration rate instead of 100\% since some space may be inaccessible to any robots.

The experiments are conducted in a customized ROS-based simulation environment where large-scale point cloud maps represent the static world. Rather than using a generic physics engine, we implement a kinematic simulator that performs ray-casting on the point cloud to simulate 3D LiDAR sensing with an effective range of 8m. The robot team consists of two GVs and two AVs. GVs are modeled with differential-drive kinematics, constrained to the ground-level traversable surfaces extracted from the point cloud. AVs are modeled with 3D kinematic mobility, allowing them to fly at a fixed altitude of 3m to bypass ground obstacles. This setup ensures that the heterogeneous traversability and sensing constraints are accurately reflected.

We evaluate all methods in three environments (Fig.~\ref{fig:sim_res}): (i) a small \textit{Garden} with little heterogeneous terrain, (ii) a large \textit{Village} with moderate heterogeneous terrain like lawn, and (iii) a large \textit{Forest} with more heterogeneous terrain such as scattered trees.
For each map, we run five independent trials and report the average exploration time and average total path length over these 5 trials.

As summarized in Table \ref{tab:map_comparison}, our \abbrName consistently outperforms the baseline approaches (\textit{Nearest} and \textit{NBVP}) with reductions of 2.3\%-35.7\% of the exploration time and total path length over the best results among the baselines.

As shown in Fig. \ref{fig:sim_res}, our approach explores faster than the other two baselines in all maps.
As the proportion of heterogeneous regions and complexity increases, the advantage of our method becomes more obvious as shown in the Village (B).
Finally, as shown in the trajectory figures, the trajectories of the baseline approaches are longer and have more intersections than ours, which indicates our approach leads to less repeated exploration and thus shortens the exploration time.

\subsubsection{Sensitivity of Terrain Analysis}
\label{sec:exp_more_agents}
\label{sec:sensitivity_terrain}

\begin{figure}[tb]
\centering
\includegraphics[width=0.88\linewidth]{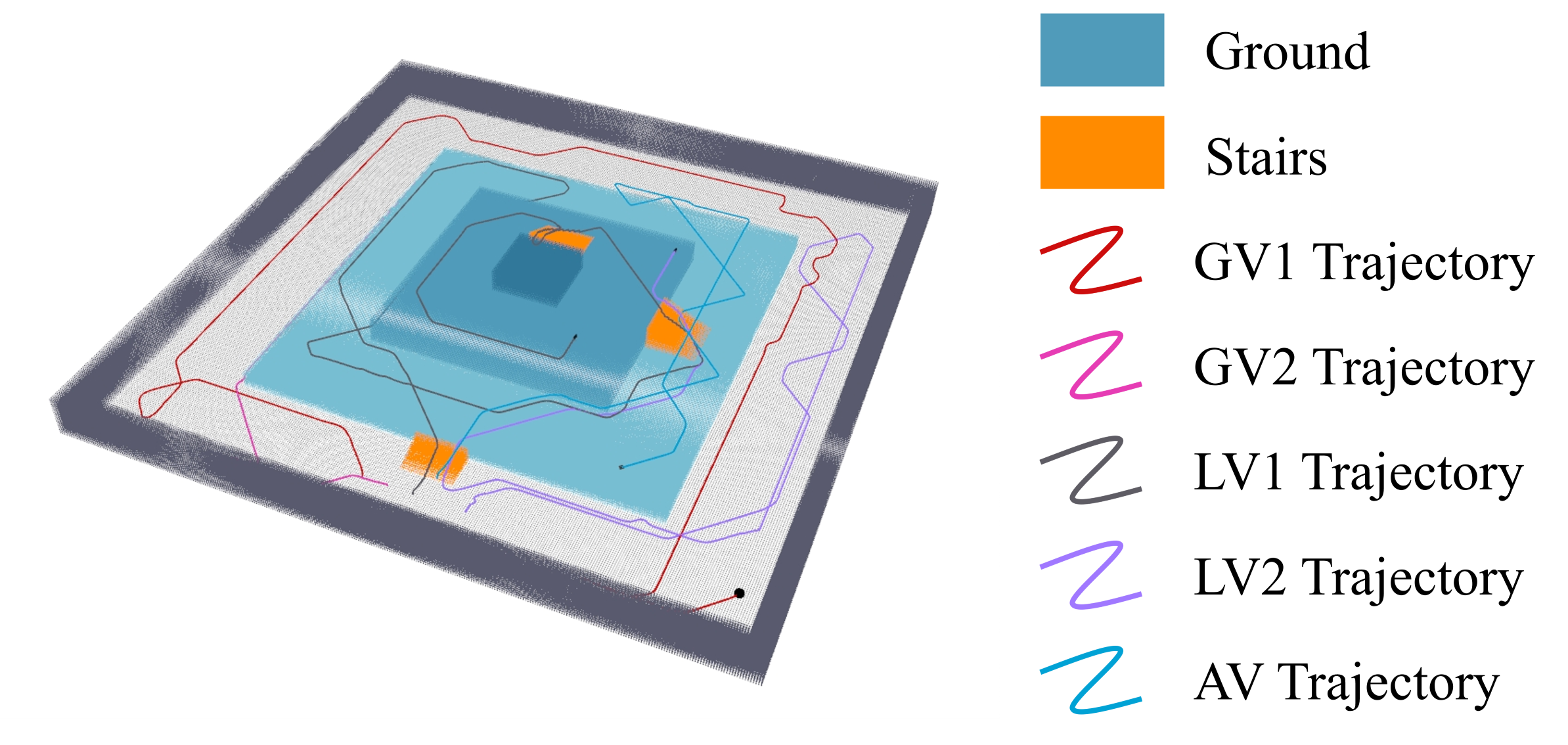}
\caption{Exploration in a four-floored environment.}
\label{fig:three-agents}
\end{figure}

We then run an experiment with five agents out of three different types with different sensing capabilities, using the same simulator described in Sec.~\ref{sec:sim_exp}.
The environment is of size $153m\times153m$ with four floors connected by staircases (Fig.~\ref{fig:three-agents}).
The team consists of two Legged Vehicles (LV), two GVs and one AV.
Two LVs share the same sensing capability ($360^\circ$ FOV) and speed limit of $0.5m/s$.
Two GVs share the same sensing range ($360^\circ$ FOV) while having different maximum speeds ($0.5m/s$ and $1m/s$).
The AV has a different FOV ($180^\circ$).
Only the LVs run terrain analysis to detect staircases to possibly go across floors.
GVs are limited to the first floor only.
The AV can only fly on the 1st and 2nd floors and is not allowed to fly on the 3rd or 4th floor (e.g. due to the height of the ceiling).

We run two experiments.
First, we vary a parameter {\texttt{surface\_tol}} in terrain analysis for stair detection used by LV.
If the height values of two neighboring cells are above this parameter, these two cells are regarded as at different floors and LV cannot traverse.
Otherwise, LV regards them as traversable.
All other parameters are fixed.
The stairs in this environment have a rise of $0.25m$ per step. 

\begin{figure}[tb]
\centering
\includegraphics[width=0.88\linewidth]{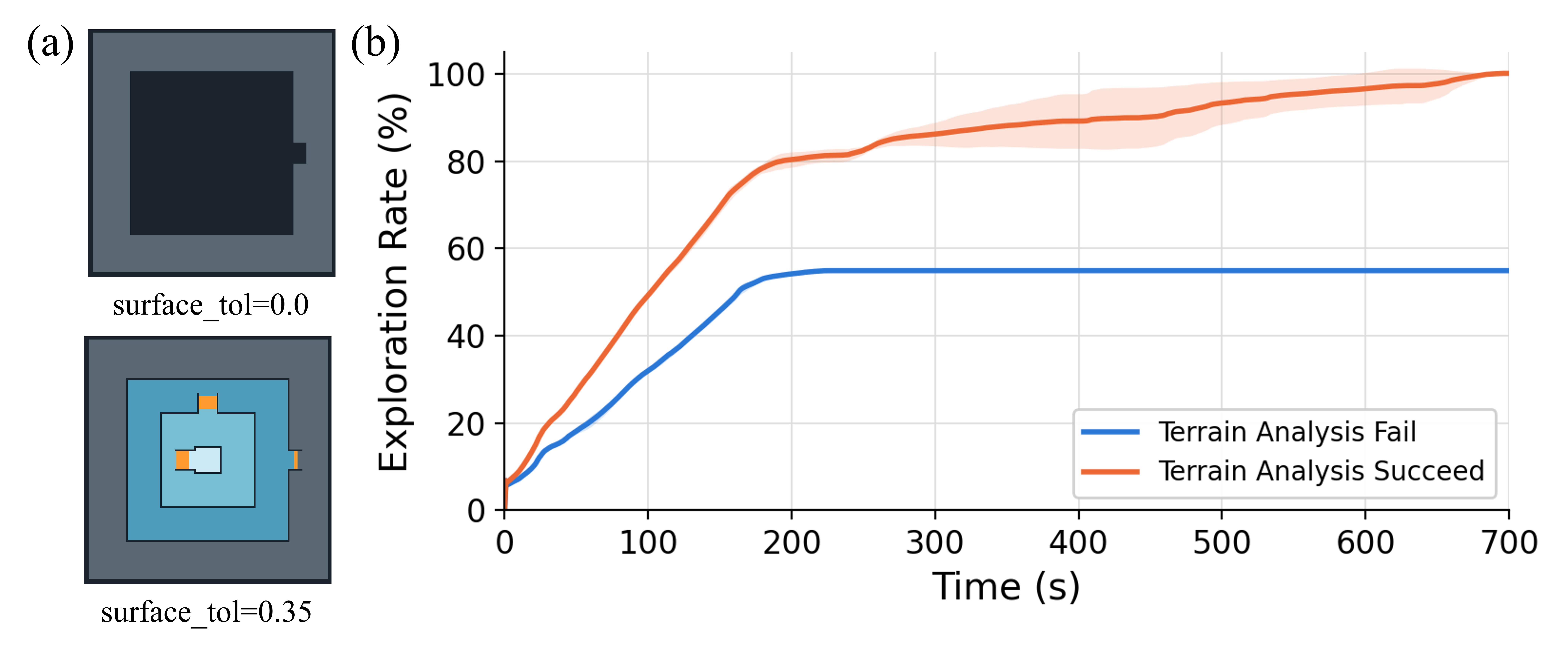}
\caption{(a) shows the terrain analysis result with varing $\texttt{surface\_tol}$. (b) shows the exploration efficiency with different terrain analysis result.}
\label{fig:terrain_result}
\end{figure}

As shown in Fig.~\ref{fig:terrain_result}(a), when {\texttt{surface\_tol}} is set to $0.0m$, below the actual step size, the stair becomes indistinguishable from a wall, and no robot can explore the 3rd and 4th floor, resulting in incomplete exploration, shown by the blue curve in Fig.~\ref{fig:terrain_result}(b).
When {\texttt{surface\_tol}} is set to $0.35$, all stairs are correctly detected by the LVs.
Our global planner can then direct the LV to the 3rd and 4th floors, resulting in complete exploration as shown by the red curve in Fig.~\ref{fig:terrain_result}(b).

\begin{figure}[tb]
    \centering
    \includegraphics[width=0.88\linewidth]{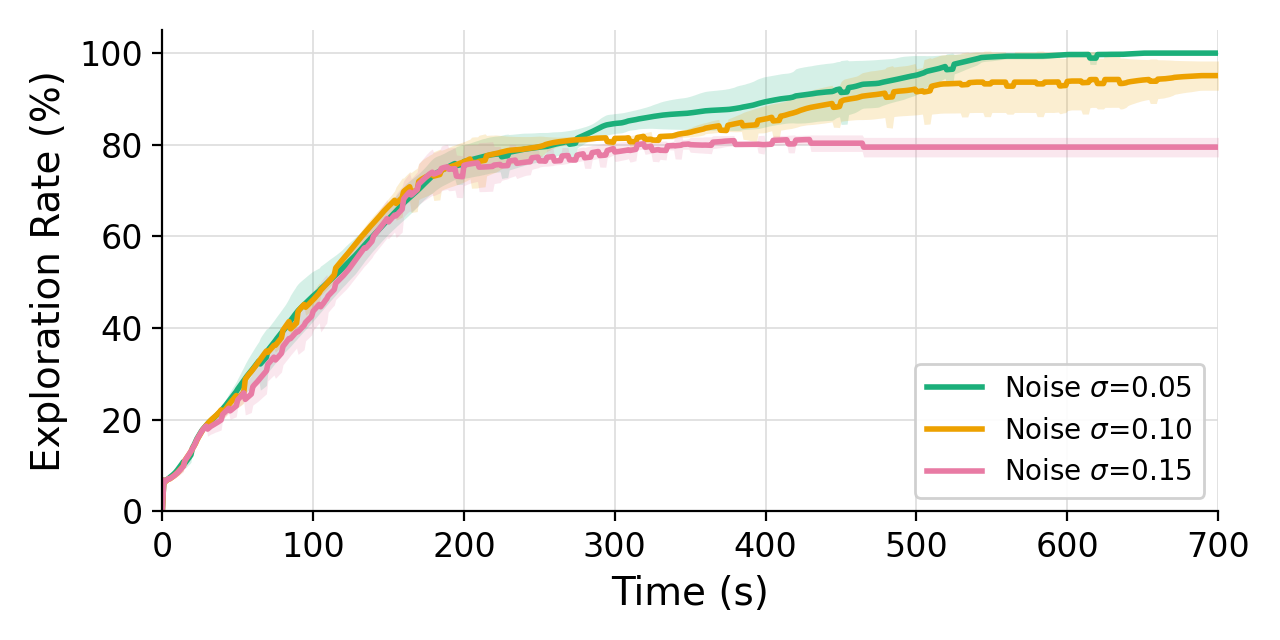}
    \caption{Exploration efficiency with different noise in terrain analysis module.}
    \label{fig:noised_terrain}
\end{figure}

The second experiment we run is to consider the uncertainty in terrain analysis.
We fix {\texttt{surface\_tol}} at $0.35m$ and perturb the estimated height difference of each cell by an additive Gaussian noise with zero mean and standard deviation $\sigma$, with $\sigma \in \{0.05, 0.10, 0.15\}m$.
With a larger $\sigma$, terrain analysis is more likely to regard stairs as walls, resulting in inaccessible space for the LVs. 
Note that, since the terrain analysis runs iteratively during exploration, whenever the height values of cells are correctly estimated in one iteration, stairs are correctly detected and LVs are directed to higher floors.
On the flip side, a correctly detected stair may be regarded inaccessible if the detection goes wrong in the next iteration.
As shown in Fig.~\ref{fig:noised_terrain}, the exploration rate decreases, as the noise in the height estimation increases (as $\sigma$ increases).
For small noise ($\sigma = 0.05m$), \abbrName still explores the whole environment. 
As $\sigma$ increases to $0.10$ and $0.15$, the exploration rates decrease to $93\%$ and $80\%$ respectively.

To summarize, with the help of the proposed planners that consider heterogeneous agents, \abbrName can resist small noise in terrain analysis by properly setting the parameters, but can result in incomplete exploration with incorrect or highly noisy terrain analysis.

\subsection{Local Planning}
\label{sec:ablation}

\begin{table}[tb]
\caption{Results with varying $\alpha$ in hetero-frontier costs.}
\label{ablation_he_reward}
\centering
\setlength{\tabcolsep}{4pt}
\begin{tabular}{l c c c}
\toprule
\textbf{Village (81m $\times$ 121m)} & \textbf{Time (s)} & \textbf{Length (m)} & \textbf{Reduction of Time} \\
\midrule
\(\alpha=0\)  & 843.7  & 1209.8 & 0\\
\(\alpha=0.2\)  & 821.3  & 1288.0 & 2.7\%\\
\(\alpha=0.4\) & 743.3  & 1189.4 & 11.9\% \\
\(\alpha=0.6\) & \textbf{704.3}  & 1189.7 & \textbf{16.5\%} \\
\(\alpha=0.8\)  & 727.0  & \textbf{1185.2} & 13.8\%\\
\(\alpha=1\) & 781.3  & 1222.1 & 7.4\% \\
\bottomrule
\end{tabular}
\end{table}

\begin{table}[tb]
\caption{Ablation Study of Local Planning. NoPr means no priority assignment. NoHe means no Hetero-frontier cost. NoLo means neither of the two tricks is included. Full means using both.}
\label{Ablation Results}
\centering
\begin{tabular}{l c c c c}
\toprule
\textbf{Map and Metric} & \textit{NoPr} & \textit{NoHe} & \textit{NoLo} & \textit{Full} \\
\midrule
\textbf{Village (81m $\times$ 121m)} & & & \\
Time Avg (s) & 758.0 & 843.7 & 896.7 & \textbf{704.3} \\
Length Avg (m) & 1398.6 & 1209.8 & 1405.3 & \textbf{1189.7}  \\
\bottomrule
\end{tabular}
\end{table}

We run an ablation study to verify the two proposed techniques in local planning.
\subsubsection{Hetero-Frontier Cost}
To evaluate the impact of hetero-frontier cost in our local planning, we vary the parameter $\alpha$ in Eq.~\ref{heha:eq:hetero_cost} from $0$ to $1$ with a step size of $0.2$, where $\alpha=0$ means no use of the proposed hetero-frontier cost, and compare the exploration time taken by our \abbrName in a large \textit{village} environment (Fig.~\ref{fig:exploration results}B(ii)), which is explained later.

As shown in Table~\ref{ablation_he_reward}, with an intermediate $\alpha = 0.6$, \abbrName achieves the shortest exploration time while $\alpha=0.8$ produces the shortest path.
A moderate $\alpha$ can direct the capable robot to those hetero-frontiers more proactively, which helps avoid having this robot take a detour to go back to those hetero-frontiers at the end.
In addition, too large an $\alpha$ (e.g. $\alpha=1$) can lead to myopic and greedy behaviour, which sometimes results in zig-zag paths, and thus prolongs the exploration time and path length.

\subsubsection{Priority Assignment}

To evaluate the effectiveness of priority assignment and hetero-frontier cost, we compare \abbrName (\textit{Full}) against \abbrName without priority assignment (\textit{NoPr}), \abbrName without hetero-frontier cost (\textit{NoHe}), and \abbrName without both tricks (\textit{NoLo}). We fix $\alpha=0.6$ in this test.
As shown in Table~\ref{Ablation Results}, \abbrName has the shortest exploration time and path length, demonstrating the benefits of both proposed tricks for local planning.

\subsection{Demonstration on Real Robots}
\label{sec:real-world_experiment}
\begin{figure*}[t]
    \centering
    \includegraphics[width=0.88\linewidth]{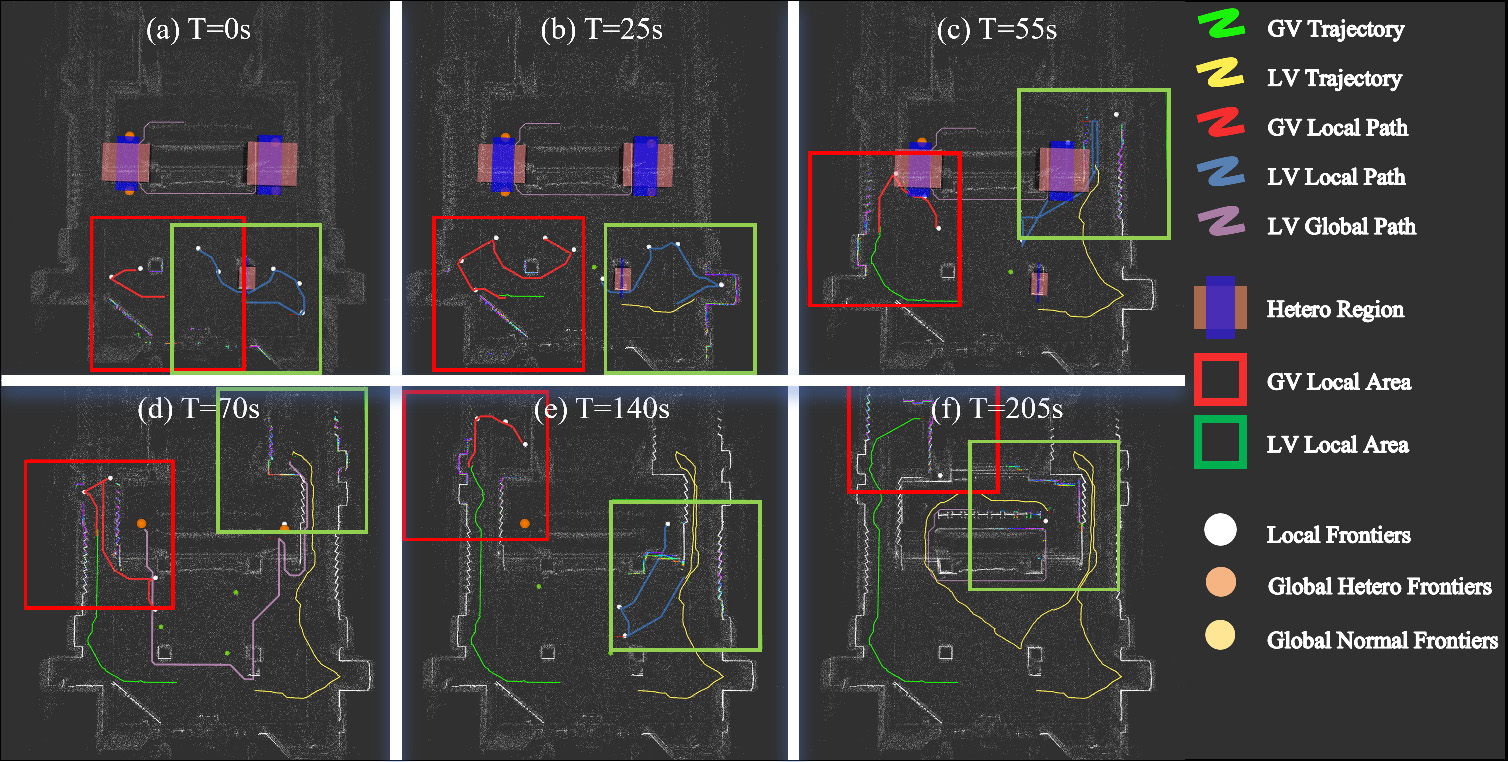}
    \caption{
        Real-world exploration using \abbrName. 
        The green and yellow lines show the executed trajectories of the ground vehicle (GV) and the legged vehicle (LV) respectively.
        The red and blue lines represent the local path of the GV and LV respectively.
        The red and green boxes indicate the local planning range of the GV and LV respectively. 
        The violet line denotes the global path of the LV.
        Due to the size of the space, GV only explores locally and there is no global path for the GV.
        The white points represent the robots’ local viewpoints, the orange points represent global heterogeneous points, and the green points represent global normal viewpoints.
        The hetero-regions are the areas with stairs where only LV can traverse.
    }
    \label{fig:expmap}
\end{figure*}

As shown in Fig.~\ref{fig:exploration results} and Fig.~\ref{fig:expmap}, we deploy our \abbrName system to explore a space of size around 35m$\times$35m with both flat ground and stairs.
We use a legged robot (Unitree Go2) and a wheeled robot (AgileX Scout Mini), both equipped with a NUC13 i7-1360P and a Livox MID360 Lidar.
We used FAST-LIO2~\cite{fastlio2} for localization and mapping on both robots.
We set the LiDAR field of view to 7 meters with 360 degrees coverage, and the local exploration range was configured as a 10-meter square.
The maximum speed of all robots was set to 0.7 m/s.
The exploration finishes in about 205 seconds, with the process shown in the snapshots in Fig.~\ref{fig:expmap}.
In (a–c), both robots initiate exploration from their starting poses on the ground level. During this stage, the local planner coordinates the robots' trajectories to ensure efficient coverage, as shown by the green line and yellow line.

As shown in (d), upon encountering a dead end, the legged robot is redirected (shown by the violet line) by the global planner toward unexplored areas, as illustrated in (e). Finally in (f), the legged vehicle ascends the stairs and stops at the second floor.

\section{Conclusion and Future Work}

\subsection{Conclusion}
This paper considers the exploration of an unknown environment using multiple heterogeneous robots such as drones, wheeled, and legged robots, which have different capabilities to traverse complex terrains.
We propose \abbrName which includes a new routing algorithm \abbrOurMTSP for its global planning that can quickly find bounded sub-optimal solutions for mHPP to minimize the maximum path length among the agents subject to traversability constraints.
Additionally, the local planner in \abbrName also considers heterogeneity to avoid repeated and duplicated exploration among the robots.
The experimental results verify the advantage of \abbrOurMTSP over existing approaches for mHPP, and that our \abbrName can reduce the exploration time by up to 30\% compared with the baseline methods

\subsection{Future Work}
For future work, one can consider developing a distributed version of \abbrOurMTSP that allows running on various robots, and extending \abbrName to its decentralized counterpart to handle communication-denied environments.
A possible way to handle communication limit is to run \abbrName on each robot independently, which is possible as the entire system does not require much computational resources as verified in our experiments, and then fuse the subset of robots together when these robots are within the communication range of each other.
When fusing these robots together, the one among them with the strongest computational resources can be selected as the leading robot, which then runs \abbrOurMTSP to plan paths for all robots, which is then dispatched to the other robots.
This is possible due to the fast running speed of \abbrOurMTSP and that it requires only modest computational resources (e.g. without any requirement on powerful GPU).

\bibliographystyle{IEEEtran}
\bibliography{refs}

\begin{IEEEbiography}[{\includegraphics[width=1in,height=1.25in,clip,keepaspectratio]{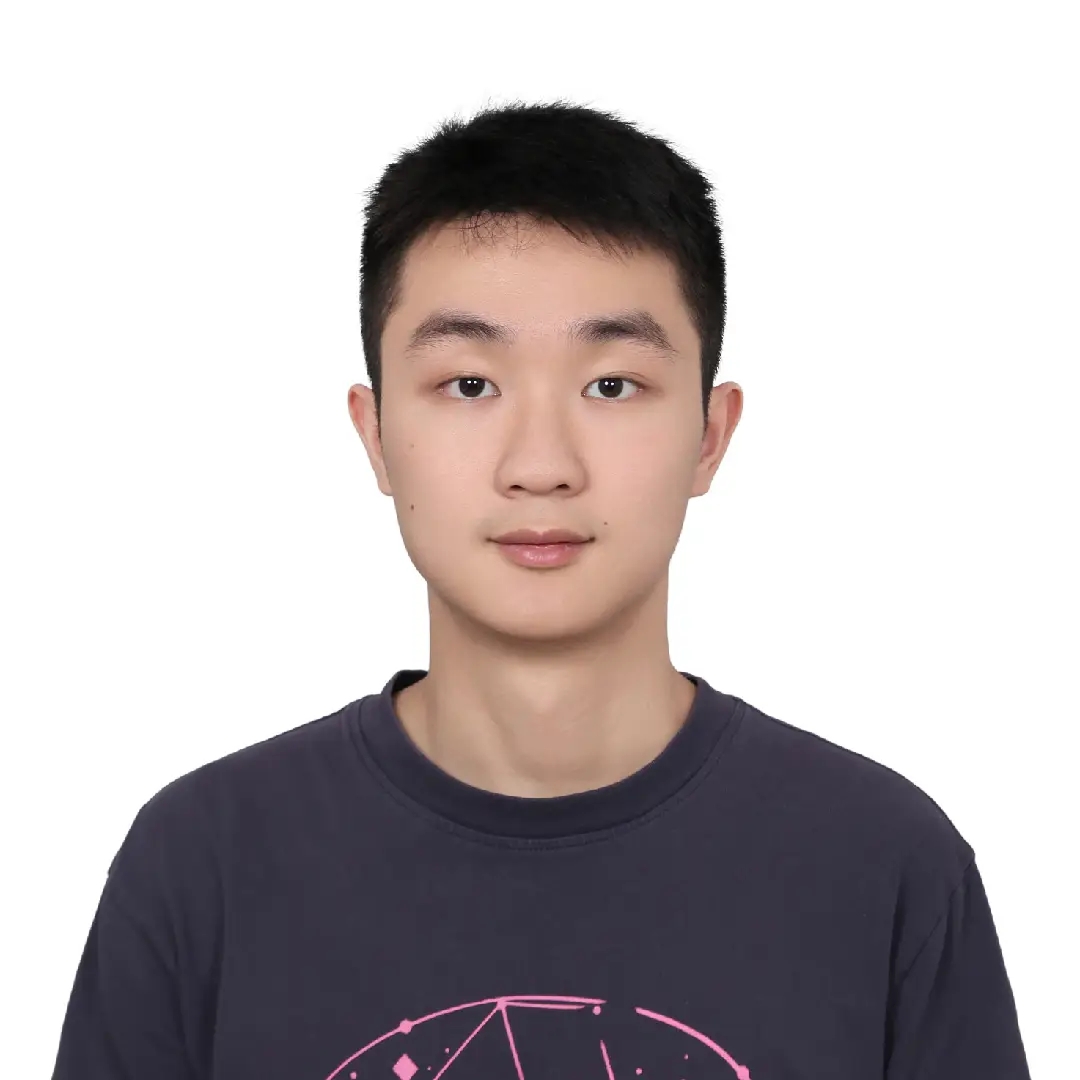}}]
{Longrui Yang}
    (Student Member, IEEE) received the B.S. degree from Lanzhou University of Technology, Lanzhou, China, in 2024. He is currently pursuing the master's degree at Nanyang Technological University (NTU), Singapore. His research interests include robot planning, autonomous exploration, embodied intelligence.
\end{IEEEbiography}

\begin{IEEEbiography}[{\includegraphics[width=1in,height=1.25in,clip,keepaspectratio]{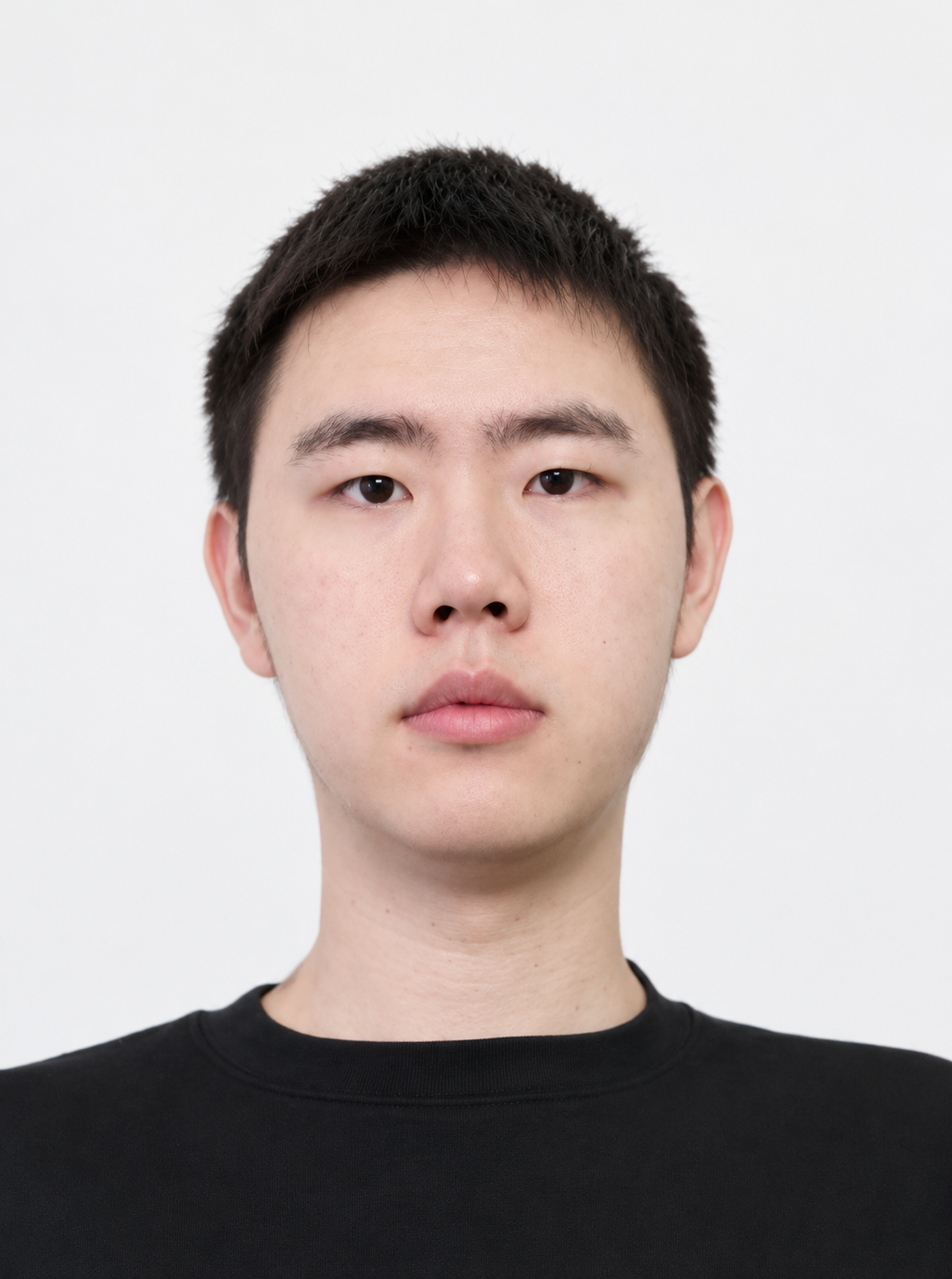}}]
{Yiyu Wang}
    (Student Member, IEEE) is working toward bachelor’s degrees in mechanical engineering at Shanghai Jiao Tong University, Shanghai, China, and in computer science at the University of Michigan, Ann Arbor, MI, USA. 
    His research interests include embodied artificial intelligence, agentic robotics, and planning and search algorithms.
\end{IEEEbiography}

\begin{IEEEbiography}[{\includegraphics[width=1in,height=1.25in,clip,keepaspectratio]{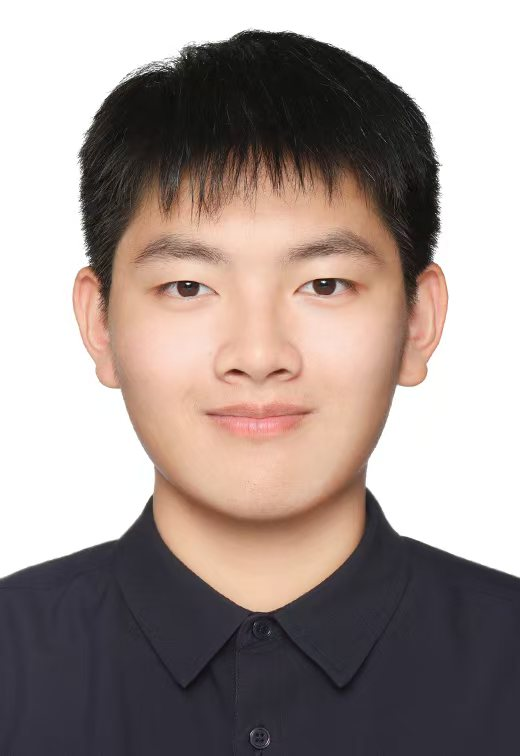}}]
    {Jingfan Tang}
    (Student Member, IEEE) is currently working toward the B.Eng. degree in electrical and computer engineering with Shanghai Jiao Tong University, Shanghai, China.
    His research interests include embodied artificial intelligence, world models, and planning and search algorithms.
\end{IEEEbiography}

\begin{IEEEbiography}[{\includegraphics[width=1in,height=1.25in,clip,keepaspectratio]{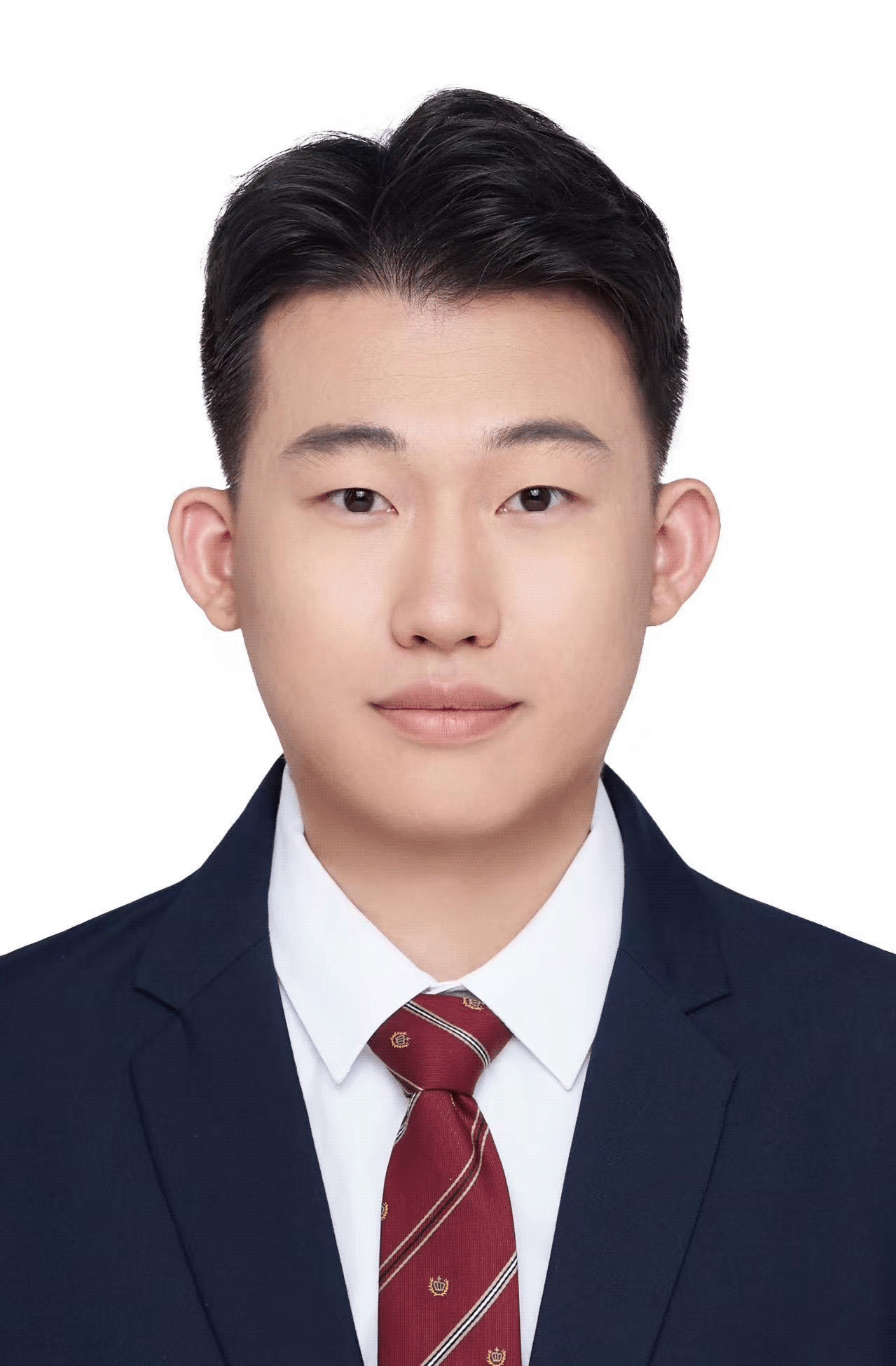}}]
    {Yunpeng Lyu} received the B.S. degree in mechanical engineering from Huazhong University of Science and Technology, Wuhan, China, in 2024.
    He is currently working toward the Ph.D. degree in control science and engineering with Shanghai Jiao Tong University, Shanghai, China.
    His research interests include robotics, target search, and embodied artificial intelligence.
\end{IEEEbiography}

\begin{IEEEbiography}[{\includegraphics[width=1in,height=1.25in,clip,keepaspectratio]{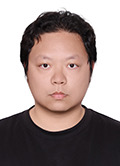}}]
{Shizhe Zhao} (Member, IEEE) received the B.S. degree from Xidian University in 2014, and the M.S. and Ph.D. degrees from Monash University in 2018 and 2023, respectively. He is currently a postdoctoral researcher at Shanghai Jiao Tong University. His research interests include heuristic search and multi-agent motion planning.

\end{IEEEbiography}

\begin{IEEEbiography}[{\includegraphics[width=1in,height=1.25in,clip,keepaspectratio]{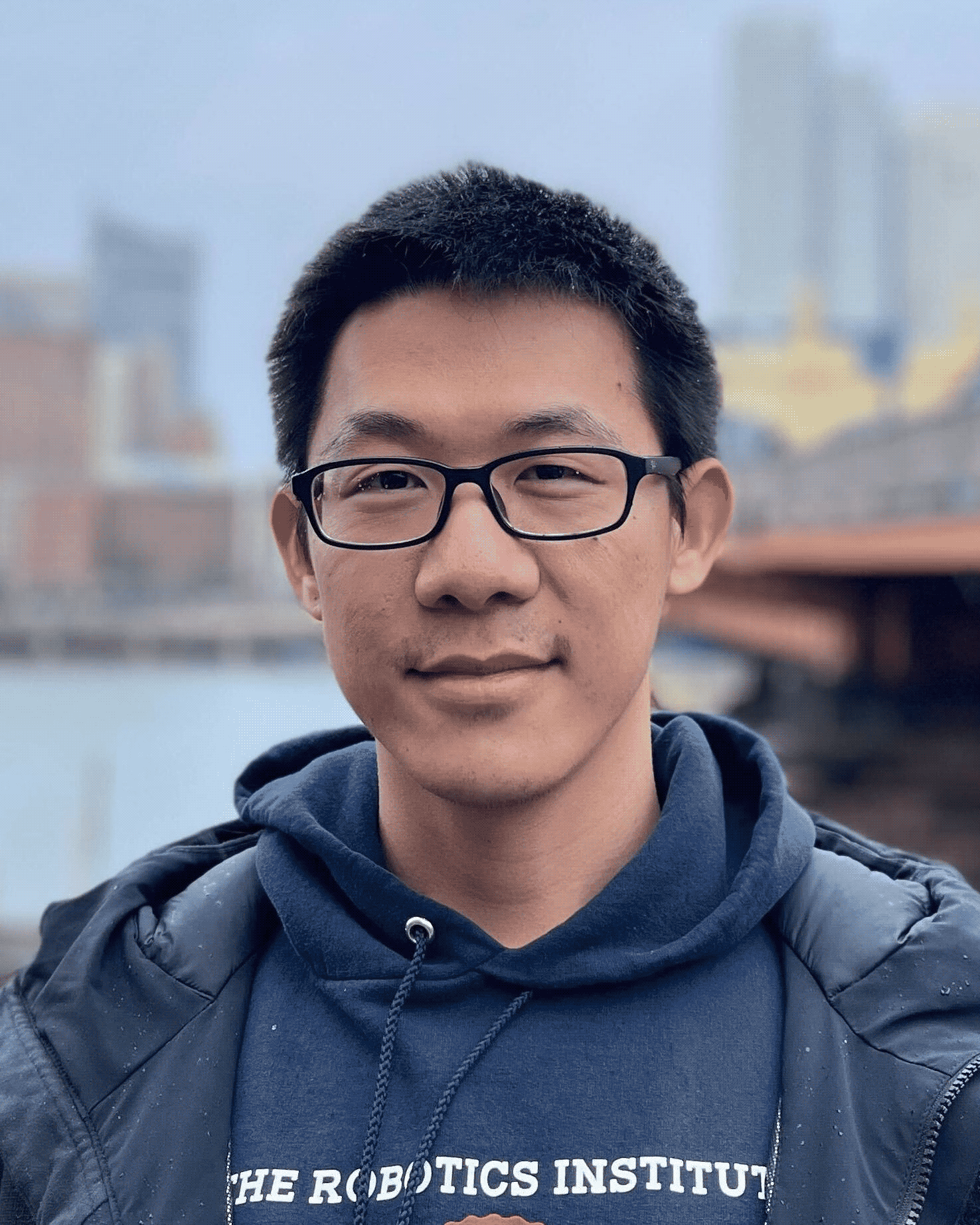}}]
    {Chao Cao} (Member, IEEE) received the B.S. degree in Computer Science from The University of Hong Kong, Hong Kong, in 2017, and the M.S. degree in robotics from the Robotics Institute, Carnegie Mellon University, Pittsburgh, PA, USA, in 2019. He is currently working toward the Ph.D. degree in robotics with the Robotics Institute, Carnegie Mellon University.
    His research interests include path and motion planning, mobile manipulation, autonomous exploration, and multi-robot coordination.
\end{IEEEbiography}

\begin{IEEEbiography}[{\includegraphics[width=1in,height=1.25in,clip,keepaspectratio]{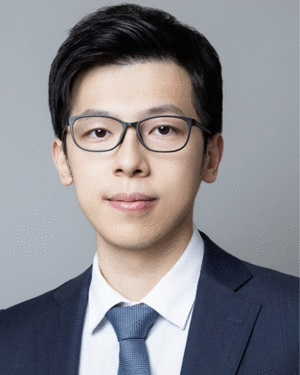}}]
    {Zhongqiang (Richard) Ren} (Member, IEEE) received the dual B.E. degree in mechatronics from Tongji University, Shanghai, China, and FH Aachen University of Applied Sciences, Aachen, Germany, in 2015, and the M.S. degree in mechanical engineering and the Ph.D. degree in mechanical engineering from Carnegie Mellon University, Pittsburgh, PA, USA, in 2017 and 2022, respectively.
    He is currently an Associate Professor with the Global College, Shanghai Jiao Tong University, Shanghai, China.
    His research interests include robotics, path and motion planning, and multi-robot systems.
    He is an Associate Editor for IEEE Transactions on Robotics, Autonomous Robots, ICRA and IROS.
\end{IEEEbiography}

\end{document}